\documentclass{article}
\PassOptionsToPackage{numbers}{natbib}
\usepackage[preprint]{neurips_2026}

\usepackage[utf8]{inputenc}
\usepackage[T1]{fontenc}
\usepackage{hyperref}
\usepackage{url}
\usepackage{xurl}
\usepackage{booktabs}
\usepackage{amsmath}
\usepackage{amsfonts}
\usepackage{amssymb}
\usepackage{microtype}
\usepackage[table]{xcolor}
\usepackage{graphicx}
\usepackage{subcaption}
\usepackage{array}
\usepackage{fvextra}
\usepackage{multirow}

\RecustomVerbatimEnvironment{verbatim}{Verbatim}{breaklines,breakanywhere,fontsize=\small}

\definecolor{TableHeader}{RGB}{232,240,254}
\definecolor{TableStripe}{RGB}{247,249,252}
\definecolor{TableHighlight}{RGB}{232,245,233}
\definecolor{PlanBlue}{RGB}{225,236,250}
\definecolor{RealizeGreen}{RGB}{229,244,234}
\definecolor{SafeguardAmber}{RGB}{255,244,220}
\definecolor{YesGreen}{RGB}{36,122,72}
\definecolor{PartAmber}{RGB}{184,120,0}
\definecolor{NoGray}{RGB}{120,120,120}

\newcommand{\Yes}{\textcolor{YesGreen}{\checkmark}}
\newcommand{\Part}{\textcolor{PartAmber}{$\sim$}}
\newcommand{\No}{\textcolor{NoGray}{--}}
\newcommand{\ModelLogo}[1]{\raisebox{-0.16em}{\includegraphics[height=0.95em]{#1}}\hspace{0.3em}}

\title{AnchorSIPS: A Synthetic Dataset and Evaluation Resource for Evidence-Supported Psychosis-Risk Symptom Measurement}
\author{%
\textbf{Guilherme C. Oliveira$^{1,*}$ \quad
Stephanie Fong$^{1,2}$ \quad
Zimu Wang$^{1,4}$ \quad
Clarice Lee$^{2}$} \\
\textbf{Xiangyu Zhao$^{1}$ \quad
Duy Khoa Pham$^{1,3}$ \quad
Duong Nhu$^{1}$ \quad
Yiwen Jiang$^{1}$ \quad
Jiahe Liu$^{1}$} \\
\textbf{Zhongxing Xu$^{1}$ \quad
Dwarikanath Mahapatra$^{5}$ \quad
Dominic Dwyer$^{1,2,*}$ \quad
Zongyuan Ge$^{1,*}$} \\
$^{1}$ AIM for Health Lab, Monash University\\
$^{2}$ Orygen and The University of Melbourne \\
$^{3}$ Swinburne University of Technology, Melbourne \\
$^{4}$ University of Liverpool \\
$^{5}$ Khalifa University \\
\texttt{\{zongyuan.ge\}@monash.edu}, \\ \texttt{dominic.dwyer@orygen.org.au}
}

\begin{document}
\maketitle

\begin{abstract}
    Progress on AI for psychosis-risk assessment is limited by a data-access bottleneck. Real clinical interviews are difficult to share because of privacy, governance, and consent constraints. We present AnchorSIPS, a synthetic dataset of 10K structured psychosis-risk interviews with transcript-grounded measurement targets. Each interview is modeled on Mini-SIPS, a clinician-administered psychosis-risk interview. It captures history, 24 symptom questions, follow-up evidence for items the patient affirms, decisions about delusion-like symptoms (unusual beliefs), hallucination-like symptoms (unusual perceptions), and disorganized communication, exclusion of clear psychotic-level symptoms ("frank psychosis"), and a final attenuated psychosis syndrome (APS) diagnosis, a high-risk state of milder or early psychotic symptoms. The APS diagnosis is not a standalone label. It depends on earlier endorsements, supporting follow-up details, symptom-class decisions, and the frank-psychosis check. Every intermediate decision is anchored to its supporting transcript turns. AnchorSIPS is generated by a plan-then-realize pipeline. A hidden case sheet specifies the patient's clinical state, a deterministic planner fixes the interview structure, and an LLM realizes only the patient utterances under validation and bounded repair. Fixing labels and structure before generation avoids the inter-turn inconsistencies typical of multi-turn LLM dialogue. Across seven LLM baselines, models recover coarse decisions but fail to extract follow-up details or cite supporting transcript turns, so final-label performance overstates interview competence. AnchorSIPS is intended for research on evidence extraction, transcript-grounded measurement, and uncertainty under partial disclosure.
\end{abstract}

\section{Introduction}

Psychosis is a mental state in which perception, beliefs, or thought organization can become disconnected from shared reality. Common forms include hallucinations, delusions, and disorganized communication. Progress on AI for psychosis-risk assessment is constrained by a data-access bottleneck: authentic clinical interviews are difficult to share, because of privacy, governance, and consent constraints in high-sensitivity psychiatric care. This is especially limiting for structured psychosis-risk assessments such as Mini-SIPS, a brief clinician-administered interview~\citep{woods2024psychs}. Mini-SIPS does not produce a single diagnostic label. Instead, the clinician moves through history and symptom queries, targeted follow-up, and class-level decisions about delusion-like symptoms (unusual beliefs), hallucination-like symptoms (unusual perceptions), and disorganized communication. They then exclude frank psychosis (clear psychotic-level symptoms) before reaching a final high-risk decision~\citep{miller2003sips,yung2005caarms}. Publicly reusable resources rarely preserve this structure. Without it, we cannot tell whether a model recovers the intermediate clinical state that a correct final judgment depends on. Synthetic data offers a principled way around the access barrier, but recent synthetic clinical-dialogue resources mostly target note generation, augmentation, or broad psychiatric diagnosis---not psychosis-risk workflow measurement~\citep{wang2024notechat,wu2024callm,psycotalk2026}.

The need for structural fidelity is sharpened by partial disclosure. In psychosis-risk interviews, clinically relevant information often emerges gradually. Stigma, disclosure concerns, and uncertainty about unusual experiences make young people at risk reluctant to describe symptoms directly~\citep{corrigan2004stigma,clement2015stigma,rusch2013helpseeking,rietdijk2011pathways}. Patients may be guarded, vague, downplay symptoms, or appear inconsistent across turns. This is exactly the kind of underspecified, multi-turn dialogue in which strong LLMs commit to early answers and fail to recover from a wrong turn~\citep{laban2025llmslost}. A trustworthy model needs to do more than predict the final label. It should identify transcript-supported symptom cues, normalize them into follow-up fields such as frequency, distress, and functional impact, and stay uncertain when the dialogue does not justify a stronger conclusion~\citep{tam2024quest,asgari2025creola}. Otherwise, final-label accuracy can overstate true interview competence~\citep{maynez2020faithfulness,rashkin2023attribution}.

We present \textit{AnchorSIPS}, a synthetic clinically aligned resource for evidence-supported psychosis-risk symptom measurement, consisting of 10{,}000 synthetic structured-interview bundles. Each bundle contains an intake and history block, 24 symptom queries with follow-up evidence for endorsed items, symptom-class decisions for delusion-like, hallucination-like, and disorganized-communication symptoms, a frank-psychosis exclusion decision, and a final attenuated psychosis syndrome (APS) diagnosis. APS is a high-risk state involving milder or early psychotic symptoms. Crucially, every decision is paired with the specific transcript turns that support it. Prior synthetic psychiatric dialogue resources mainly target diagnosis, note generation, or augmentation. AnchorSIPS instead makes the structured-interview workflow itself the prediction target. Models are scored on whether they recover each intermediate decision and cite the transcript turns that justify it. The resource supports research on transcript-grounded extraction, intermediate clinical reasoning, evidence citation, and uncertainty under partial disclosure.

Our main contributions are
\begin{enumerate}
	\item \textbf{A workflow-aligned synthetic resource paired with transcript turn IDs.} AnchorSIPS makes the structured interview workflow---query endorsements, follow-up fields, class-level decisions, frank-psychosis exclusion, and APS diagnosis---the prediction target. Each target is linked to the specific transcript turns that support it. To our knowledge, this is the first synthetic psychosis-risk resource to release field-level transcript evidence alongside workflow labels.
	\item \textbf{An auditable plan-then-realize pipeline that restricts the LLM to surface realization.} All workflow labels are derived deterministically from a hidden case sheet during planning; the LLM only writes patient-response wording under validation and bounded repair. This makes released labels independent of the generator and avoids the inter-turn drift of end-to-end dialogue generation.
	\item \textbf{A baseline evaluation that quantifies the gap between coarse decisions and grounded measurement.} Across seven LLM baselines, models recover coarse workflow decisions more reliably than detailed follow-up fields or transcript-linked evidence. These results identify evidence extraction, grounded citation, and intermediate workflow consistency as key bottlenecks for psychosis-risk NLP systems.
\end{enumerate}

\section{Related Work}

\paragraph{Structured psychosis-risk assessment and psychosis-risk NLP.}
AnchorSIPS is motivated by structured psychosis-risk interviews such as SIPS, CAARMS, Mini-SIPS, and PSYCHS, which operationalize psychosis-risk assessment through guided elicitation and staged clinical judgment~\citep{miller2003sips,yung2005caarms,woods2024psychs,bilgrami2025ampscz}. It is not presented as a protocol-equivalent clinical simulator. It is also complementary to psychosis-risk NLP on real interview data: prior work links language features with psychosis risk and clinician-rated symptoms~\citep{morgan2021nlpmarkers,bilgrami2022constructvalidity}, while CHiRPE studies real semi-structured psychosis-risk interviews with clinician-oriented explanation formats~\citep{fong2026chirpe}. Relative to this literature, AnchorSIPS contributes a controllable synthetic benchmark for transcript-grounded symptom measurement and intermediate workflow recovery.

\paragraph{Clinical dialogue datasets and synthetic clinical conversations.}
Synthetic health data is increasingly used to address privacy, governance, and data-access barriers in clinical research~\citep{beaulieujones2019privacy,gonzales2023synthetic,smolyak2024synthetichealth,miletic2024tabularhealth}. Open clinical-dialogue resources such as MTS-Dialog and ACI-Bench support note generation from encounter transcripts~\citep{benabacha2023mtsdialog,yim2023acibench}, while recent LLM-based work studies synthetic clinical conversations for augmentation, reconstruction, and evaluation, including NoteChat, SynDial, CliniChat, CALLM, and PsyCoTalk~\citep{wang2024notechat,das2024synthetic,chen2025clinichat,wu2024callm,psycotalk2026}. A parallel line of work develops counseling and mental-health dialogue resources, including CPsyCoun, Cactus, Crisp, D4, MentalChat16K, MDD-5k, and DiagESC~\citep{zhang-etal-2024-cpsycoun,lee-etal-2024-cactus,zhou-etal-2025-crisp,yao2022d4,shenlab2025mentalchat16k,yin2025mdd5k,seo-lee-2024-diagesc}. These resources are valuable, but are mainly note-centric, support-centric, diagnosis-centric, or counseling-centric rather than focused on structured psychosis-risk symptom measurement. AnchorSIPS instead releases workflow-aligned psychosis-risk interview targets with transcript-linked evidence. Table~\ref{tab:dataset-comparison} summarizes adjacent resources along the properties needed for this benchmark: psychosis-risk specificity, interview or dialogue text, workflow-aligned intermediate targets, transcript-linked evidence, and labels fixed independently of patient-utterance realization.

\begin{table}[htbp]
	\centering
	\caption{Comparison of AnchorSIPS with related clinical-dialogue, counseling, mental-health, and psychosis-risk resources. \Yes = yes; \Part = partial; \No = no. ``Labels fixed'' indicates that benchmark labels are specified independently before patient-utterance realization.}
	\label{tab:dataset-comparison}
	\scriptsize
	\setlength{\tabcolsep}{3.8pt}
	\renewcommand{\arraystretch}{1.18}
	{\rowcolors{3}{TableStripe}{white}
	\resizebox{\textwidth}{!}{%
	\begin{tabular}{p{0.21\linewidth}p{0.16\linewidth}p{0.14\linewidth}p{0.16\linewidth}cccc}
		\rowcolor{TableHeader}
		\toprule
		\textbf{Resource} & \textbf{Primary focus} & \textbf{Text object} & \textbf{Construction} & \textbf{Psychosis-risk} & \textbf{Workflow targets} & \textbf{Turn evidence} & \textbf{Labels fixed} \\
		\midrule
		\rowcolor{TableHeader}\multicolumn{8}{@{}l}{\textbf{Psychosis-risk and clinical speech resources}} \\
		AMP SCZ / PSYCHS resource~\citep{bilgrami2025ampscz} & psychosis-risk cohort & language/speech tasks & real cohort data & \Yes & \Part & \No & \No \\
		SPEAK study~\citep{bayer2023speak} & thought disorder / psychosis & spoken interviews & real cohort data & \Yes & \Part & \No & \No \\
		\addlinespace[0.15em]
		\rowcolor{TableHeader}\multicolumn{8}{@{}l}{\textbf{General clinical-dialogue and interview reconstruction resources}} \\
		ACI-Bench~\citep{yim2023acibench} & visit-note generation & clinical encounters & real and synthetic & \No & \Part & \Part & \No \\
		MTS-Dialog~\citep{benabacha2023mtsdialog} & note generation & doctor-patient dialogues & real transcripts & \No & \No & \No & \No \\
		MEDIQA-Chat 2023~\citep{benabacha2023mediqachat} & summarization / generation & doctor-patient dialogues & mixed task data & \No & \No & \No & \No \\
		NoteChat~\citep{wang2024notechat} & note-conditioned dialogue & patient-physician dialogues & synthetic from notes & \No & \No & \No & \Part \\
		SynDial~\citep{das2024synthetic} & note-conditioned dialogue & patient-physician dialogues & synthetic from notes & \No & \No & \No & \Part \\
		CliniChat / MedQA-Dialog~\citep{chen2025clinichat} & clinical interview ability & reconstructed interviews & knowledge-driven synthetic & \No & \Part & \No & \Part \\
		\addlinespace[0.15em]
		\rowcolor{TableHeader}\multicolumn{8}{@{}l}{\textbf{Mental-health counseling, support, and diagnosis-dialogue resources}} \\
		CPsyCoun~\citep{zhang-etal-2024-cpsycoun} & psychological counseling & counseling dialogues & report-based reconstruction & \No & \Part & \No & \Part \\
		Cactus~\citep{lee-etal-2024-cactus} & CBT counseling & counseling dialogues & LLM / role simulation & \No & \Part & \No & \No \\
		Crisp~\citep{zhou-etal-2025-crisp} & cognitive restructuring & supportive dialogues & LLM distilled & \No & \Part & \No & \No \\
		D4~\citep{yao2022d4} & depression diagnosis & doctor-patient dialogues & simulated clinical sessions & \No & \Part & \No & \Part \\
		MentalChat16K~\citep{shenlab2025mentalchat16k} & mental-health assistance & counseling / support chats & synthetic and anonymized & \No & \Part & \No & \No \\
		MDD-5k~\citep{yin2025mdd5k} & mental-disorder diagnosis & diagnostic dialogues & neuro-symbolic synthesis & \No & \Part & \No & \Part \\
		DiagESC / DESC~\citep{seo-lee-2024-diagesc} & depression + support & emotional-support dialogues & synthetic with filtering & \No & \Part & \No & \Part \\
		CALLM~\citep{wu2024callm} & clinical interview analysis & mental-health interviews & LLM augmentation & \No & \Part & \No & \Part \\
		PsyCoTalk~\citep{psycotalk2026} & psychiatric comorbidity & diagnostic dialogues & synthetic from records & \No & \Part & \No & \Part \\
		\rowcolor{TableHighlight}
		\textbf{AnchorSIPS (Our work)} & \textbf{psychosis-risk measurement} & \textbf{structured interviews} & \textbf{plan-then-realize synthetic} & \Yes & \Yes & \Yes & \Yes \\
		\bottomrule
	\end{tabular}
	}
	}
	\renewcommand{\arraystretch}{1.0}
\end{table}

The comparison highlights the core contribution: AnchorSIPS jointly provides psychosis-risk specificity, structured interview text, workflow-aligned targets, transcript-linked evidence, and an auditable generation pipeline in which workflow labels are fixed before patient utterances are realized.

\paragraph{Evidence-grounded evaluation in high-stakes NLP.}
AnchorSIPS also connects to evidence-grounded evaluation in high-stakes NLP. Recent healthcare LLM work argues that systems should be assessed not only by final-answer plausibility, but also by safety, uncertainty handling, and trustworthy evidence use~\citep{tam2024quest,asgari2025creola}. More general work on faithfulness and attribution similarly shows that fluent outputs often overstate what is actually supported by the underlying evidence~\citep{maynez2020faithfulness,honovich2022true,rashkin2023attribution,li2024attributionbench}. These concerns are especially important in psychosis-risk interviewing, where clinically relevant information may emerge gradually, ambiguously, or inconsistently across turns. AnchorSIPS operationalizes this issue by making transcript-linked evidence, intermediate workflow state, and underdetermined cases explicitly measurable.

\section{Dataset Construction}
AnchorSIPS is a synthetic dataset and evaluation resource for psychosis-risk symptom measurement, organized around an observable interview workflow rather than proposed as a new psychosis-risk instrument. The synthetic design addresses privacy and governance barriers to sharing authentic psychosis-risk interviews~\citep{beaulieujones2019privacy,gonzales2023synthetic}. The transcript-linked targets reflect the need to evaluate high-stakes NLP systems for evidence support rather than final-answer plausibility alone~\citep{tam2024quest,rashkin2023attribution}.

\begin{figure}[!htb]
	\centering
	\includegraphics[scale=0.25]{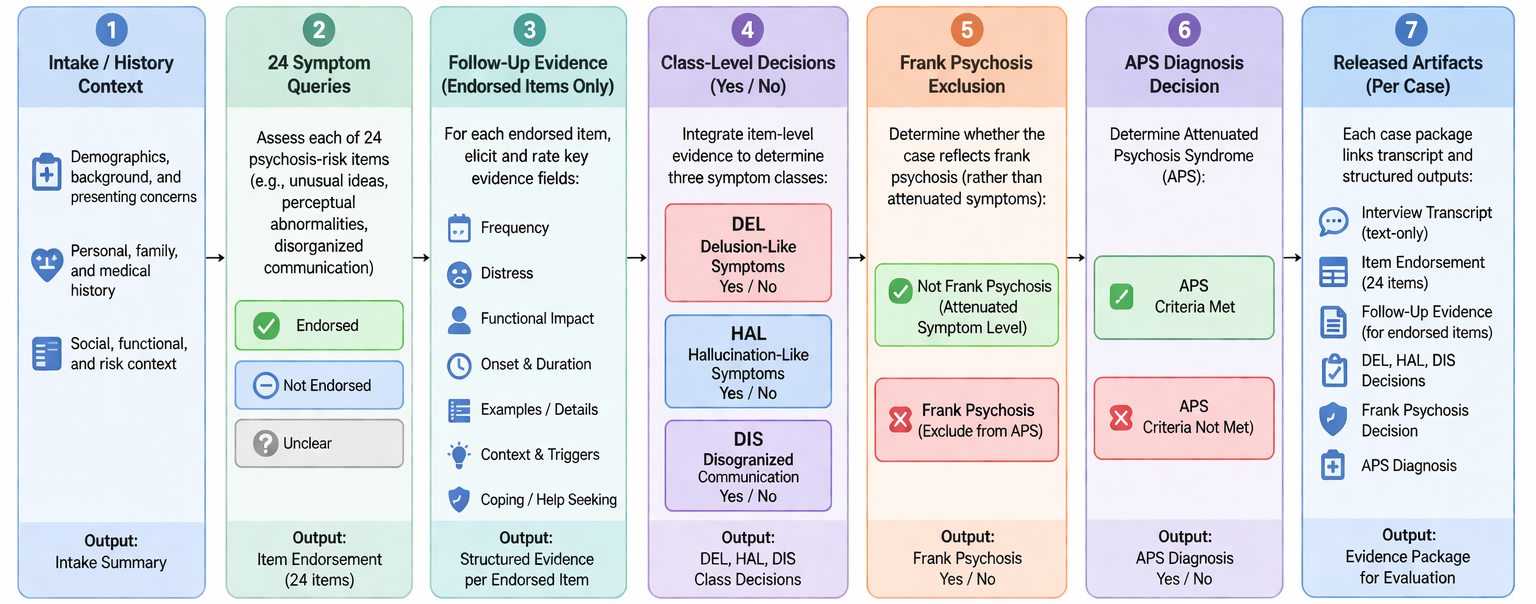}
	\caption{AnchorSIPS is framed around evidence-supported psychosis-risk symptom measurement, operationalized through the observable interview workflow rather than the internal synthetic generator. The primary outputs are transcript-linked measurement targets.}
	\label{fig:pipeline}
\end{figure}

\paragraph{Instrument Alignment}
AnchorSIPS operationalizes the observable workflow of a structured psychosis-risk interview as symptom-measurement targets(Fig~\ref{fig:pipeline}). This workflow is important because the final APS diagnosis is not a standalone label. It depends on earlier evidence collected during the interview, including which symptom questions are endorsed, what follow-up details support those endorsements, whether symptom-class criteria are met, and whether frank psychosis is excluded. Representing these steps makes it possible to evaluate not only whether a model predicts the final decision, but also whether it recovers the information needed to justify that decision from the transcript. Each interview bundle therefore contains intake and history context, responses to 24 symptom queries, follow-up evidence for endorsed items, class-level decisions for delusion-like symptoms (DEL), hallucination-like symptoms (HAL), and disorganized communication (DIS), a frank-psychosis exclusion decision, and a final attenuated psychosis syndrome (APS) diagnosis. Here, \emph{frank psychosis exclusion} refers to whether the presentation is better described as psychotic-level symptoms rather than attenuated psychosis-risk symptoms. The 24-query backbone and follow-up domains were adapted from the public Mini-SIPS form and cross-checked against SIPS, CAARMS, and PSYCHS descriptions; wording-level deviations are treated as resource-specific adaptations rather than verbatim instrument reproduction~\citep{minisips2020form,miller2003sips,yung2005caarms,woods2024psychs}.

\subsection{Plan-then-Realize Generation}
We adopt a plan-then-realize design because AnchorSIPS is a measurement benchmark, not only a source of fluent synthetic dialogue. One-shot prompting and multi-agent role-play can generate natural clinical-looking conversations~\citep{wang2024notechat,psycotalk2026}, but they entangle hidden-case specification, interview policy, patient language, and label assignment in a single generation step. For workflow-aligned evaluation, this is risky because skipped queries, off-policy follow-up, premature diagnostic conclusions, or symptom drift can silently become label noise. The structured-interview setting makes deterministic planning especially appropriate because the final APS decision depends on observable intermediate steps, including fixed symptom queries, follow-up evidence, class-level DEL/HAL/DIS criteria, and frank-psychosis exclusion, rather than on unconstrained conversation alone. Deterministic planning therefore encodes the interview coverage and decision dependencies, while the LLM is used only where flexibility is needed, namely patient-language variation.

\begin{table}[htbp]
	\centering
	\caption{Plan-then-realize pipeline. AnchorSIPS fixes the case state, released labels, and interview structure before any patient text is generated. The LLM is restricted to patient-utterance realization, while validation and repair enforce alignment with the fixed plan.}
	\label{tab:plan-then-realize}
	\footnotesize
	\setlength{\tabcolsep}{3.5pt}
	\renewcommand{\arraystretch}{1.18}
	{\rowcolors{3}{TableStripe}{white}
	\begin{tabular}{@{}p{0.13\linewidth}p{0.22\linewidth}p{0.43\linewidth}p{0.14\linewidth}@{}}
		\rowcolor{TableHeader}
		\toprule
		\textbf{Phase} & \textbf{Step} & \textbf{Description} & \textbf{Actor} \\
		\midrule
		\cellcolor{PlanBlue}\textbf{Planning} &
		1. Latent ground truth &
		Hidden case sheet defines demographics, symptom states, confounds, disclosure controls, and all released workflow labels. &
		Deterministic \\

		\cellcolor{PlanBlue}\textbf{Planning} &
		2. Interview plan &
		Fixed trajectory with intake/history, all 24 symptom queries, endorsed-item follow-ups, interviewer text, and turn metadata. &
		Deterministic \\

		\cellcolor{RealizeGreen}\textbf{Realization} &
		3. Patient wording &
		Patient utterances are generated one turn at a time from the planned target meaning, patient profile, and recent context. &
		LLM \\

		\cellcolor{SafeguardAmber}\textbf{Safeguard} &
		4. Validation &
		Checks structure, metadata consistency, denial polarity, truncation, and local plan-text compatibility. &
		Deterministic \\

		\cellcolor{SafeguardAmber}\textbf{Safeguard} &
		5. Bounded repair &
		Only flagged patient turns may be changed. Interviewer turns, unaffected context, and released labels remain fixed. &
		Deterministic/LLM \\
		\bottomrule
	\end{tabular}
	}
	\renewcommand{\arraystretch}{1.0}
\end{table}

This design is also motivated by evidence that LLMs are unreliable in underspecified multi-turn settings. \citet{laban2025llmslost} show that strong LLMs often make early assumptions, attempt final answers prematurely, and fail to recover after a wrong conversational turn. AnchorSIPS therefore separates content planning from surface realization, following the standard NLG distinction~\citep{reiter2000building} and using agenda-style disclosure controls for partial observability~\citep{schatzmann2007agenda}. Concretely, a hidden case sheet defines the intended presentation, a deterministic planner fixes the interview backbone, query coverage, follow-up policy, and disclosure schedule, and the LLM is restricted to writing the wording of one patient utterance at a time. In the released generation run, this patient-utterance realization step used GPT-OSS-120B. Two properties follow from this restriction: (i) all released workflow labels---query endorsements, follow-up fields, class-level decisions, frank-psychosis exclusion, and APS diagnosis---are derived deterministically during planning, before any patient text exists; and (ii) the LLM has no authority over interview structure, follow-up policy, or label assignment. The trade-off is less unconstrained conversational freedom, but greater auditability, reproducibility, and cleaner transcript-grounded evaluation.
Table~\ref{tab:plan-then-realize} summarizes the key design boundary. Planning fixes what the interview means, while realization controls only how the patient says it. Appendix~\ref{app:algorithm} gives the full deterministic generation sequence, and Appendix~\ref{app:planner} details how the interview plan is constructed from the fixed case sheet.

\section{Dataset Release and Composition}
\begin{figure}[htbp]
	\centering
	\includegraphics[width=\linewidth]{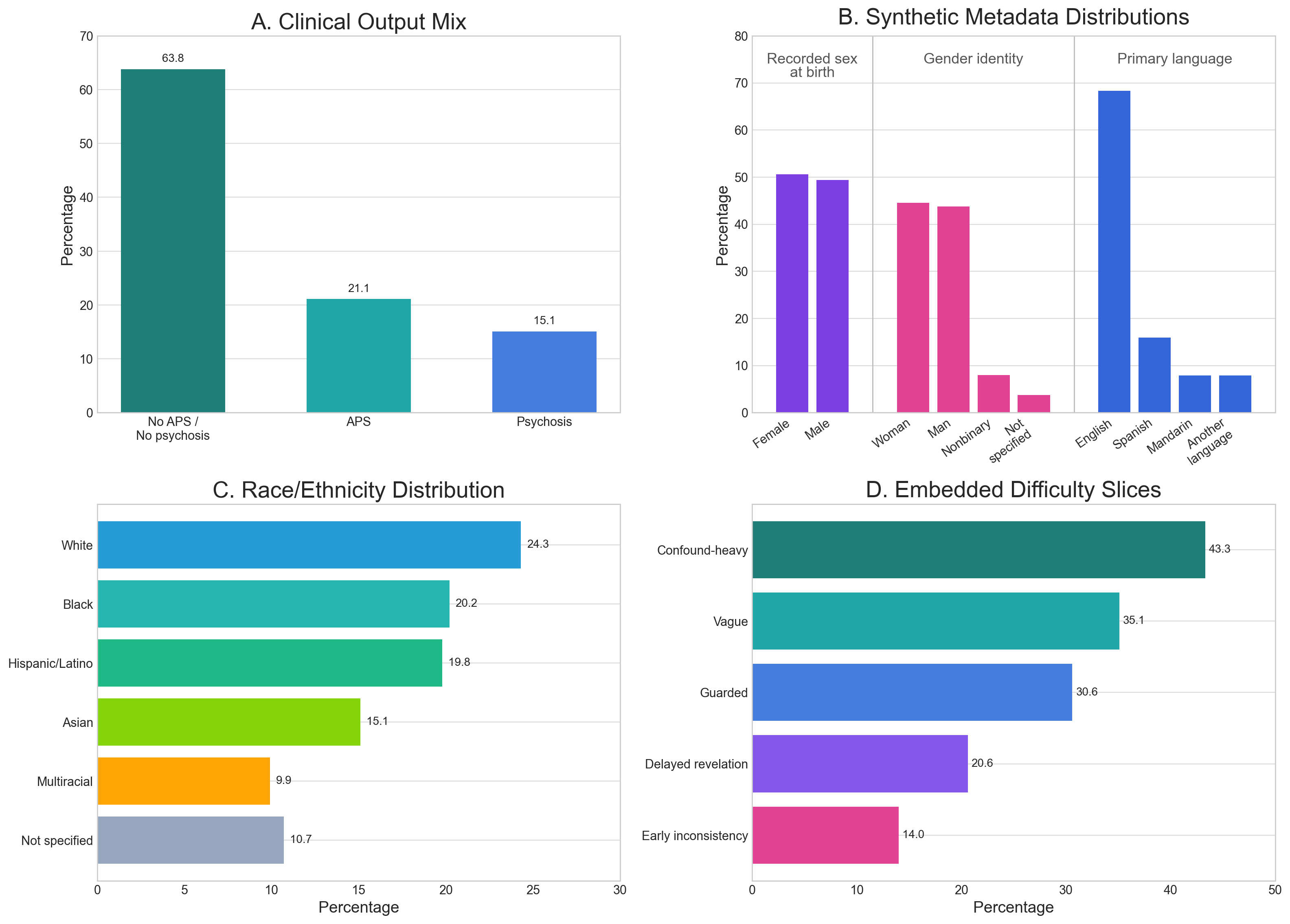}
\caption{Frozen AnchorSIPS release composition. Panel A shows the released clinical output mix. Panels B and C summarize surfaced synthetic metadata distributions. Panel D reports frequencies of embedded difficulty slices.}
	\label{fig:composition}
\end{figure}

AnchorSIPS comprises 10{,}000 synthetic interviews. The clinical output mix is 63.8\% no-APS/no-psychosis, 21.1\% APS, and 15.1\% psychosis. These proportions are intended to resemble referral-enriched psychosis-risk assessment settings rather than direct population prevalence estimates~\citep{fusar2016darkside,fusar2016pretest,nelson2007uhrreferral}. Demographic and context variables are sampled from predefined distributions rather than from a single clinical catchment or a strict balancing scheme. These variables are included as synthetic metadata for evaluation stratification and subgroup stress testing, not as estimates of population prevalence or as causal clinical predictors. In the surfaced release metadata, recorded sex assigned at birth is 50.6\% female and 49.4\% male. Gender identity is 44.5\% woman, 43.8\% man, 8.0\% nonbinary, and 3.7\% not specified. Primary language is 68.3\% English, 15.9\% Spanish, 7.9\% Mandarin, and 7.9\% another language. Race/ethnicity label is 24.3\% White, 20.2\% Black, 19.8\% Hispanic/Latino, 15.1\% Asian, 9.9\% Multiracial, and 10.7\% not specified. Figure~\ref{fig:composition} summarizes these distributions.

Difficulty characteristics are embedded within the main interview-level release rather than separated into special splits. In the frozen release, 30.6\% of interviews are Guarded, 35.1\% Vague, 20.6\% show strong delayed revelation, 14.0\% show early inconsistency, and 43.3\% are confound-heavy. These properties define reusable evaluation slices for downstream stress testing while preserving the interview bundle as the primary release unit. The same interviews support multiple task views, including query endorsement, follow-up evidence extraction, class-level DEL/HAL/DIS decisions, and final diagnostic decisions. 

\section{Expert Evidence-Support Audit}

We used three psychosis-risk interview-trained research reviewers to conduct an expert audit of label-evidence alignment in the released synthetic interviews. The reviewers had a combined 12+ years of relevant experience in youth mental health, psychosis-focused research, clinical research assessment, and structured interviewing, including PSYCHS interviews with participants at ultra-high risk for psychosis and community controls and hundreds of in-depth interviews with young people about psychotic-like experiences. In structured psychosis-risk interviews, trained researchers also score symptom criteria from participant responses, making their expertise appropriate for this evidence-support audit.

The audit asks whether benchmark-facing class-level APS-range decision records are understandable, traceable, and supported by the observable transcript snippets provided to reviewers. This is an internal validity check for the benchmark: it evaluates whether released labels are supported within the synthetic interview evidence, but does not establish that AnchorSIPS labels correspond to diagnoses in real patients or that model performance on AnchorSIPS implies deployment readiness. The audit rubric (Table~\ref{tab:human-rubric-main}) is therefore a task-specific, literature-informed instrument rather than a formally validated psychometric scale. Its dimensions are chosen to align with prior work on healthcare LLM evaluation and evidence grounding, which emphasizes evaluator expertise, explicit task criteria, uncertainty, and the distinction between plausible outputs and outputs that are actually supported by the available evidence \citep{tam2024quest,maynez2020faithfulness,honovich2022true,rashkin2023attribution,li2024attributionbench}.

\begin{table}[!htp]
    \centering
    \caption{Expert audit rubric for criterion-level evidence support.}
    \label{tab:human-rubric-main}
    \small
    \setlength{\tabcolsep}{4pt}
    \renewcommand{\arraystretch}{1.12}
    {\rowcolors{2}{TableStripe}{white}
    \begin{tabular}{@{}p{0.27\linewidth}p{0.68\linewidth}@{}}
        \rowcolor{TableHeader}
        \toprule
        \textbf{Criterion} & \textbf{Question} \\
        \midrule
        Symptom presence & Does the participant describe the target experience or behaviour? \\
        Current weekly occurrence & Does the participant say it happens currently and at least weekly? \\
        Worsening in past year & Does the participant say it began or became worse in the past year? \\
        Distress & Does the participant say it is upsetting, frustrating, or distressing? \\
        Functional influence & Does the participant describe an effect on communication, relationships, work, study, or daily life? \\
        Alternative explanation / rule-out & Does the evidence support the intended rule-out status, including whether stress, poor sleep, substances, anxiety, or another explanation is present or reasonably excluded? \\
        \bottomrule
    \end{tabular}
    }
    \renewcommand{\arraystretch}{1.0}
\end{table}

\begin{figure}[!htp]
	\centering
	\includegraphics[width=\linewidth]{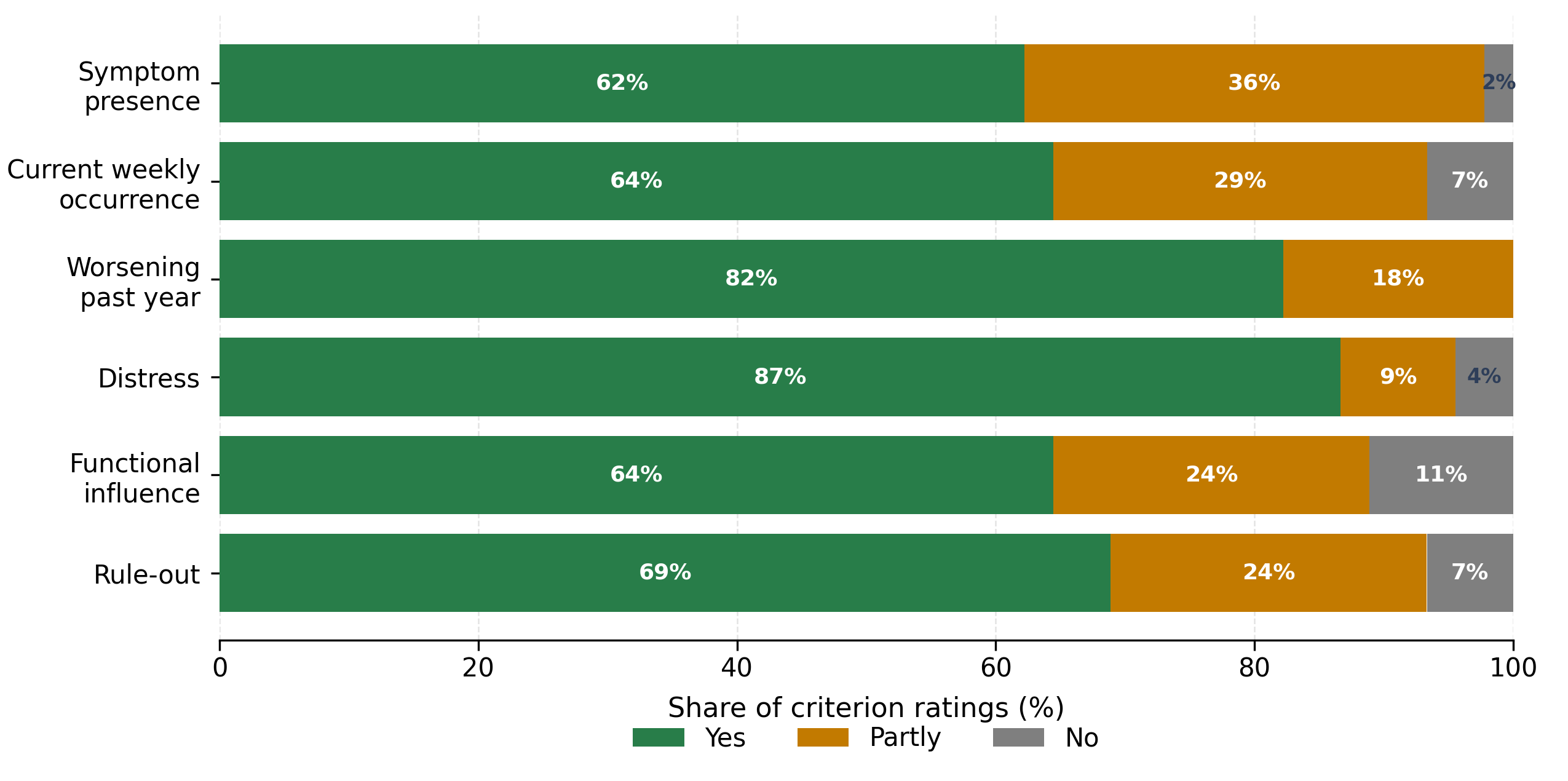}
	\caption{Expert evidence-support audit results under the six-criterion rubric. Bars show the percentage of Yes, Partly, and No ratings for each criterion across 15 interview-by-symptom-class decision records independently rated by three reviewers (45 item-level reviews; 270 criterion-level ratings).}
	\label{fig:human-review}
\end{figure}

The current audit targets class-level APS-range decision records, where each record corresponds to one interview-by-symptom-class decision for delusion-like symptoms (DEL), hallucination-like symptoms (HAL), or disorganized communication (DIS). Each record is extracted from frozen benchmark-facing workflow bundles and sampled from the frozen benchmark shard with balancing over symptom class, diagnosis label, and representative disclosure-difficulty slices. Each record is presented as an email-contained audit package with the symptom area, the benchmark-facing automated decision, and criterion-labeled transcript snippets. A representative reviewer email package is shown in Appendix~\ref{app:expert-audit-email}. Reviewers rate whether the observable transcript evidence supports six criterion-level dimensions: symptom presence, current weekly occurrence, worsening in the past year, distress, functional influence, and alternative explanation / rule-out. All criteria are rated on the same three-point scale: Yes, Partly, or No.

Figure~\ref{fig:human-review} summarizes the completed audit, which checks whether the evidence shown to reviewers supports the intended labels. We evaluated 15 interview-by-symptom-class decision records, each rated independently by all three reviewers, yielding 45 item-level reviews and 270 criterion-level ratings. Overall, reviewers usually judged the cited transcript snippets as supporting the intended criterion-level labels. The clearest evidence was for whether the symptom had worsened in the past year and whether it caused distress. Ratings were less decisive for functional influence and alternative-explanation/rule-out judgments, consistent with the greater ambiguity of inferring impairment and differential explanations from brief transcript excerpts. Pairwise exact agreement across item-criterion ratings was 57.0\%, with a mean absolute ordinal difference of 0.51 on the No--Partly--Yes scale. These results support the internal evidence-label alignment of the reviewed AnchorSIPS records, while identifying functioning and rule-out judgments as important targets for larger future audits.

\section{Evaluation Tasks and Baseline Experiments}
AnchorSIPS evaluates whether models can recover workflow-aligned symptom judgments from full interview transcripts, rather than only predict a plausible final diagnosis. The same released interviews support linked tasks spanning query endorsement, follow-up extraction, class-level DEL/HAL/DIS decisions, and final frank-psychosis and APS decisions. We evaluate whether a model can recover both the interview form values and the transcript evidence supporting them.

\begin{itemize}
	\item \textbf{Query endorsement.} Macro positive-class F1 across the 24 symptom questions. For example, if a participant says they sometimes hear a whisper when alone, the model should mark the relevant hallucination-like or unusual-perception query as endorsed.
	\item \textbf{Follow-up extraction.} Mean token-level F1 over endorsed query-field instances. This measures whether the model recovers details such as frequency, time course, distress, functional impact, conviction or reality testing, and alternative explanations.
	\item \textbf{DEL/HAL/DIS class-level decisions.} Macro positive-class F1 across class-decision fields. This measures whether the model aggregates item-level evidence into the correct symptom area: delusion-like symptoms, hallucination-like symptoms, or disorganized communication.
	\item \textbf{Frank-psychosis exclusion.} Positive-class F1 for the decision about whether clear psychotic-level symptoms are excluded.
	\item \textbf{APS diagnosis.} Positive-class F1 for the final attenuated psychosis syndrome decision. This requires combining intermediate evidence rather than simply detecting one symptom mention.
	\item \textbf{Evidence linking.} Evidence-link F1, measuring overlap between predicted and reference supporting transcript turn IDs. In the whisper example, a good model should cite the turns where the participant described the whisper, its frequency, recent worsening, distress, and impact on sleep.
\end{itemize}

For ranking only, we report a workflow composite score (WCS), computed as the unweighted mean of five harder task metrics: follow-up field F1, DEL/HAL/DIS macro-F1, frank-psychosis F1, APS F1, and evidence-link F1. Query endorsement is reported as a primary task metric but excluded from WCS because it is near-ceiling for most systems and is less diagnostic of grounded measurement. The main results table in the body (Table~\ref{tab:main-results}) reports the four most diagnostic measurement-focused metrics directly; follow-up field F1 and WCS are reported in full in Appendix~\ref{app:extra-baseline-tables} (Table~\ref{tab:main-results-full}). Additional exact-match, grounding, citation, and consistency diagnostics are defined in Appendix~\ref{app:metric-defs} and reported in Appendix~\ref{app:extra-baseline-tables}.

\paragraph{Citation convention and metric scope.}
Evidence-link F1 reflects both substantive grounding---whether the model finds the part of the transcript that supports a decision---and adherence to the patient-turn-only citation convention encoded in the released labels. The released reference cites only the patient turn that contains the disclosure; the surrounding interviewer prompt is not annotated as supporting evidence. The baseline prompt instructs models to ``cite transcript turn IDs supporting each major section'' without specifying a single-turn convention, so a model that cites a question-and-answer turn pair receives reduced precision even when the cited region overlaps the reference. We therefore report evidence-recall as the convention-robust component of evidence-link performance, and grounded correctness (Appendix~\ref{app:metric-defs}) uses overlap-tolerant matching that is unaffected by interviewer-turn citations. Evidence-link F1 should be read as a stricter measure that combines substantive grounding with citation precision under the released convention, rather than as a pure grounding measure in isolation.

\paragraph{Baseline prompting and output parsing.}
All evaluated LLM baselines were run with a single joint form-filling prompt over the full released interview transcript rather than with separate prompts for each task. The benchmark prompt presents each turn with its released turn ID, speaker role, interview stage, symptom class, query ID, and follow-up type, and instructs the model to use only the released transcript, avoid hidden-label assumptions, and return one JSON object containing query endorsements, follow-up fields for endorsed queries, class-level DEL/HAL/DIS decisions, frank-psychosis and APS decisions, and supporting transcript turn IDs. Outputs are parsed against a strict benchmark schema, including exact required keys and exact alignment between endorsed query IDs and returned follow-up blocks; malformed outputs trigger benchmark-side JSON repair and normalization before scoring. Appendix~\ref{app:benchmark-prompts} gives the benchmark prompt template and output constraints.

Following evidence-grounded healthcare and attribution evaluations~\citep{tam2024quest,asgari2025creola,rashkin2023attribution,li2024attributionbench}, this single-pass setup is a standardized baseline rather than an optimized extraction pipeline, so results should be read as common transcript-to-JSON prompting behavior rather than as an upper bound on task-specific systems.

\subsection{Results}
Table~\ref{tab:main-results} summarizes baseline performance on the fixed 200-interview workflow-evaluation shard. The central result is a consistent gap between coarse workflow recognition and evidence-supported measurement. Query endorsement was high across evaluated systems, ranging from 0.949 to 1.000, and frank-psychosis prediction was also comparatively strong, with F1 up to 0.930. These tasks are closer to coarse recognition, where models can often succeed by detecting salient symptom mentions or clear threshold cues in the transcript. Performance dropped for structured measurement.

The hardest task in the displayed metrics was evidence linking. Evidence-link F1 remained low even for the strongest systems, topping out at 0.290, indicating that models can recover coarse decisions while citing supporting transcript turns less reliably. Follow-up field extraction shows the same pattern, ranging from 0.164 to 0.248 across the five reported models and reported in full in Appendix~\ref{app:extra-baseline-tables} (Table~\ref{tab:main-results-full}). These tasks require normalized field values and supporting transcript turns, not only a plausible clinical summary. This pattern is consistent with prior work showing that fluent model outputs can exceed their faithfulness, attribution quality, and safety-relevant evidence use~\citep{maynez2020faithfulness,honovich2022true,rashkin2023attribution,tam2024quest,asgari2025creola,li2024attributionbench}. It also matches concerns about multi-turn settings, where models can commit early and fail to recover when later evidence changes the interpretation~\citep{laban2025llmslost}. Overall, final or coarse decision accuracy can overstate interview competence when transcript-grounded measurement is evaluated directly.

\begin{table}[htbp]
	\centering
	\caption{Main baseline results on the harder workflow-aligned AnchorSIPS tasks. Higher is better; models are ordered by workflow composite score (WCS).}
	\label{tab:main-results}
	\scriptsize
	\setlength{\tabcolsep}{4.5pt}
	\renewcommand{\arraystretch}{1.16}
	{\rowcolors{3}{TableStripe}{white}
	\begin{tabular}{@{}lcccc@{}}
		\rowcolor{TableHeader}
		\toprule
		\textbf{Model}         & \textbf{DEL/HAL/DIS} & \textbf{Frank Psych.} & \textbf{APS} & \textbf{Evidence Link} \\
		\rowcolor{TableHeader}
		                       & \textbf{Macro-F1}    & \textbf{F1}           & \textbf{F1}  & \textbf{F1}            \\
		\midrule
		\ModelLogo{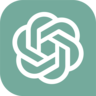}GPT-5.5                    & 0.746                & 0.930                 & 0.457        & 0.283                  \\
		\ModelLogo{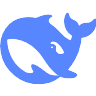}DeepSeek V4 Flash          & 0.685                & 0.930                 & 0.496        & 0.290                  \\
		\ModelLogo{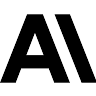}Claude Opus 4.7            & 0.728                & 0.930                 & 0.508        & 0.272                  \\
		\ModelLogo{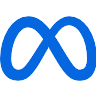}Llama 3.3 70B Instruct     & 0.655                & 0.892                 & 0.527        & 0.222                  \\
		\ModelLogo{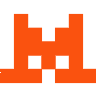}Ministral 8B 2512          & 0.692                & 0.899                 & 0.579        & 0.126                  \\
		\bottomrule
	\end{tabular}
	}

	\vspace{0.3em}
	\footnotesize{Follow-up field F1 and WCS are reported in Table~\ref{tab:main-results-full}. Two baselines (Qwen3 32B, Llama 3.1 8B Instruct) returned near-empty structured outputs under this prompt and are documented separately in Appendix~\ref{app:extra-baseline-tables}.}
\end{table}

The model ranking also shows that no evaluated model is uniformly strong. GPT-5.5, DeepSeek V4 Flash, and Claude Opus 4.7 are tightly clustered by WCS, but each remains weak on follow-up extraction and evidence linking. Ministral 8B obtains the highest APS F1 but has weak evidence-link F1, suggesting that final-label prediction can improve without reliable citation. Appendix diagnostics in Table~\ref{tab:aux-overall-results} reinforce the same conclusion. Grounded correctness remains low even for the strongest systems, evidence precision trails evidence recall, and unsupported claim rates remain nontrivial. Overall, AnchorSIPS exposes a clinically important failure mode in psychosis-risk NLP, where models can appear competent on final or coarse decisions while remaining unreliable at evidence extraction, grounding, and intermediate workflow consistency~\citep{tam2024quest,asgari2025creola,rashkin2023attribution,li2024attributionbench}.

\section{Future Directions}
Future work will focus on extensions beyond the core workflow benchmark already reported here. We plan larger frozen releases, stronger and broader human expert audits, multilingual and cultural-context stress tests, and temporal modeling across repeated interviews. We also plan a more systematic evaluation of calibration, abstention, and evidence use under partial disclosure, motivated by the current finding that coarse workflow decisions are easier for models than transcript-grounded reasoning. As with any synthetic benchmark, clinically meaningful next steps require validation on real-world interview data rather than relying on synthetic performance alone.

\section{Use Cases and Limitations}
\label{sec:usecases-limitations}

\paragraph{Intended Use Cases} AnchorSIPS is intended for methodological research on evidence-supported psychosis-risk symptom measurement rather than direct clinical use. It is designed for work on models that must read a multi-turn interview, recover structured symptom information, and justify their conclusions from transcript evidence. Three use cases are especially central: \textbf{evidence-supported symptom measurement}, where a model recovers symptom judgments supported by the dialogue; \textbf{workflow-level form completion}, where a model fills the 24 query endorsements, DEL/HAL/DIS class-level decisions, the frank-psychosis decision, and the APS diagnosis as one operationalization of that measurement problem; and \textbf{partial-disclosure stress testing}, where models are compared across guarded, vague, inconsistent, delayed-disclosure, or confound-heavy cases to assess overcalling, undercalling, or appropriate abstention when evidence is incomplete. These use cases make AnchorSIPS most appropriate for research stress testing, ablation studies, and controlled comparison of reasoning pipelines, and do not imply that good dataset performance is sufficient for safe clinical deployment.

\paragraph{Limitations and Non-Use} All reported baseline numbers reflect a single transcript-to-JSON prompting regime and should not be read as upper bounds on system performance; chain-of-thought reasoning, few-shot prompting, retrieval-augmented citation, and multi-pass extract-then-cite pipelines are not measured here and are expected to improve follow-up extraction, class-decision recovery, and evidence-link metrics. AnchorSIPS remains a synthetic dataset and evaluation resource. Patient language is generated under planner constraints rather than collected from real individuals, so linguistic patterns and conversational dynamics may differ from natural clinical interviews. Mini-SIPS itself is a clinical interview tool, and this resource does \emph{not} claim full protocol equivalence. The resource is aligned to the observable interview workflow, but its hidden planning variables are synthetic implementation choices rather than official scoring units from the instrument. Demographic and contextual fields are sampled from simplified templates and do not fully capture real sociocultural variation. Psychosis exclusion and differential explanation are especially difficult to approximate in synthetic settings, and any released labels should be interpreted as evidence-supported symptom-measurement targets rather than as diagnostic ground truth. Good performance on AnchorSIPS may be useful for methodological study and stress testing, but it is not evidence of diagnostic validity or deployment readiness. Most importantly, the dataset is \textbf{not intended for clinical diagnosis or medical decision making}; any system developed with AnchorSIPS requires validation on real-world clinical data before clinical use.

\section{Data Availability}
AnchorSIPS is released as a public research resource accompanying the manuscript release and includes synthetic interviews, workflow-aligned labels, evidence-linked justifications, benchmark evaluator code, and documentation, with official released outputs clearly separated from auxiliary internal variables. The dataset is released under CC BY-NC 4.0 with research-only terms that prohibit diagnostic, triage, treatment, and patient-facing use. The dataset is available at \url{https://huggingface.co/datasets/anonymousxxxy/resource}. The public Mini-SIPS 1.0 form is available online~\citep{minisips2020form}.

\section{Ethical Considerations}
AnchorSIPS is fully synthetic and contains no real patient transcripts, reducing but not eliminating misuse risks.  The dataset is released only for methodological research, not for diagnosis, triage, treatment planning, or patient-facing deployment. Safety controls include public release under CC BY-NC 4.0 research-only terms, prohibited-use language, and documentation that distinguishes workflow alignment from protocol equivalence and dataset performance from clinical validity. These safeguards do not remove all risk, but provide a governance baseline for responsible synthetic clinical data release.

\appendix

\section{Instrument Alignment, Fidelity, and Provenance}
\label{app:fidelity-deviations}

AnchorSIPS is workflow-aligned, not protocol-equivalent. This means the dataset follows the main visible steps of a structured psychosis-risk interview: history questions, 24 symptom queries, follow-up questions for endorsed symptoms, DEL/HAL/DIS class decisions, frank-psychosis exclusion, and final APS diagnosis. Mini-SIPS, SIPS, CAARMS, and PSYCHS are clinical or research interview frameworks for psychosis-risk assessment, but AnchorSIPS does not reproduce full clinical administration, official rater training, or clinical diagnostic validity. Instead, it uses the observable interview workflow as a benchmark structure for testing whether models can recover symptom judgments from transcript evidence. The released targets are query endorsements, follow-up fields, DEL/HAL/DIS class decisions, frank-psychosis exclusion, and APS diagnosis; hidden generator variables are used only to create synthetic cases and are not official instrument outputs.

\subsection{Question-Set Provenance}
\label{app:question-provenance}

This appendix documents the provenance of the implemented AnchorSIPS interview bank. The purpose is to distinguish item groups that are adapted from public Mini-SIPS materials from benchmark-authored prompts introduced for intake setup, dialogue flow, and controlled synthetic realization. Table~\ref{tab:question-provenance} summarizes this distinction at the item-group level.

In short, the DEL/HAL/DIS query backbone follows the implemented 24-item workflow with minor wording normalization, whereas intake wording, normalized follow-up phrasing, and dialogue scaffolds are benchmark-specific operationalizations. This distinction helps frame AnchorSIPS as workflow-aligned rather than as a verbatim or protocol-equivalent reproduction of Mini-SIPS.

\begin{table}[htbp]
	\centering
	\caption{Question-set provenance for the implemented AnchorSIPS interview bank.}
	\label{tab:question-provenance}
	\scriptsize
	\setlength{\tabcolsep}{4pt}
	\renewcommand{\arraystretch}{1.12}
	{\rowcolors{2}{TableStripe}{white}
	\begin{tabular}{p{0.18\linewidth}p{0.34\linewidth}p{0.18\linewidth}p{0.22\linewidth}}
		\rowcolor{TableHeader}
		\toprule
		\textbf{AnchorSIPS item/group} & \textbf{Current source} & \textbf{Status} & \textbf{Reason / note} \\
		\midrule
		HIST\_01--03 & Public Mini-SIPS form not used verbatim for intake wording & benchmark-specific operationalization & Short opening context prompts added for synthetic interview setup \\
		DEL\_Q01--DEL\_Q16 & Public Mini-SIPS form~\citep{minisips2020form} & minor wording adaptation & Normalize phrasing and maintain consistent benchmark wording \\
		HAL\_Q01--HAL\_Q05 & Public Mini-SIPS form~\citep{minisips2020form} & minor wording adaptation & Normalize phrasing and maintain consistent benchmark wording \\
		DIS\_Q01--DIS\_Q03 & Public Mini-SIPS form~\citep{minisips2020form} & minor wording adaptation & Normalize phrasing and maintain consistent benchmark wording \\
		Follow-up domains & Mini-SIPS / SIPS / CAARMS positive-symptom follow-up structure~\citep{miller2003sips,yung2005caarms,woods2024psychs} & instrument-derived structure & Frequency, time course, conviction or reality, distress, functioning, and alternative explanation retained as official domains \\
		Benchmark-authored prompts & AnchorSIPS implementation & benchmark-specific operationalization & Normalized follow-up wording and dialogue scaffolds for benchmark consistency and controlled disclosure \\
		\bottomrule
	\end{tabular}
	}
	\renewcommand{\arraystretch}{1.0}
\end{table}

\section{Expert Audit Email Package}
\label{app:expert-audit-email}

\begin{figure}[htbp]
	\centering
	\setlength{\fboxrule}{0.4pt}
	\setlength{\fboxsep}{1pt}
	\begin{subfigure}[t]{0.82\linewidth}
		\centering
		\fbox{\includegraphics[width=0.985\linewidth]{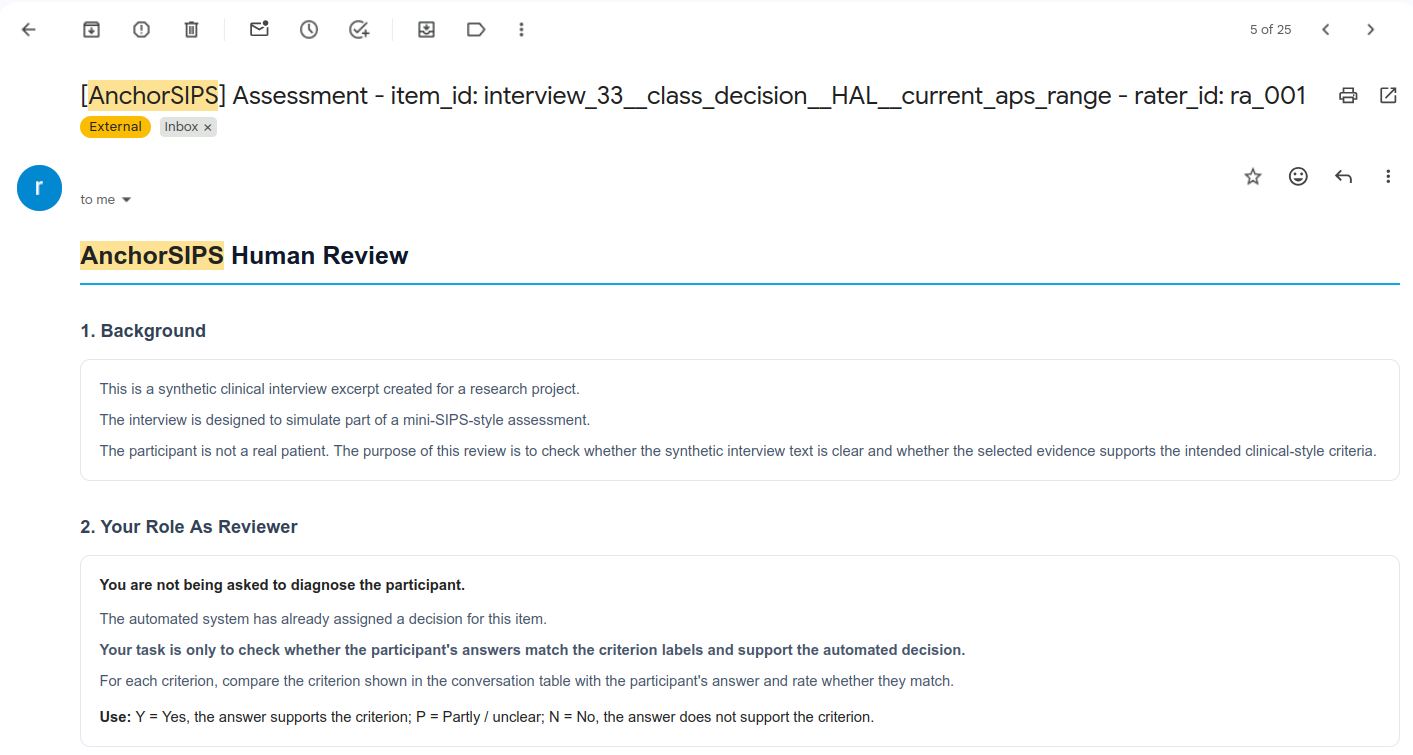}}
		\caption{Task background and reviewer role.}
	\end{subfigure}

	\vspace{0.4em}

	\begin{subfigure}[t]{0.82\linewidth}
		\centering
		\fbox{\includegraphics[width=0.985\linewidth]{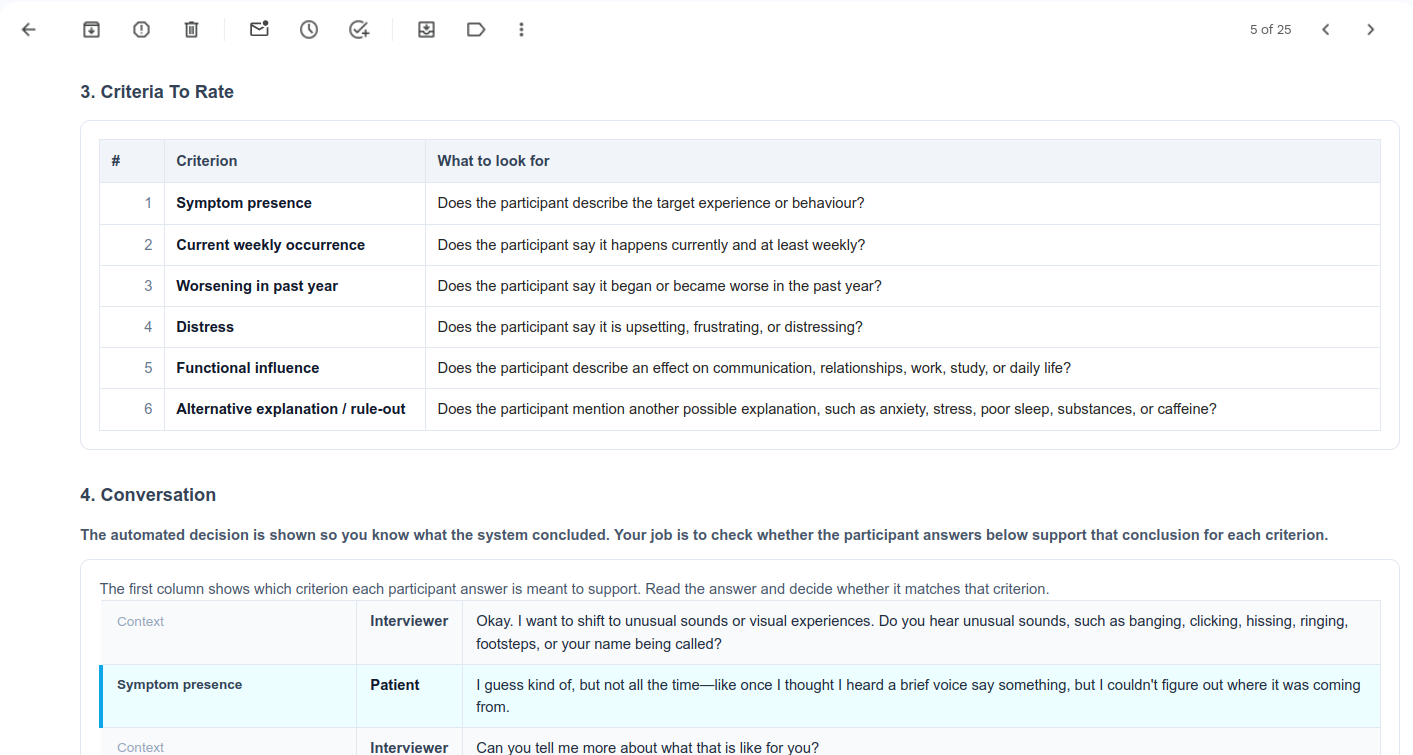}}
		\caption{Rating criteria and conversation table.}
	\end{subfigure}

	\vspace{0.4em}

	\begin{subfigure}[t]{0.82\linewidth}
		\centering
		\fbox{\includegraphics[width=0.985\linewidth]{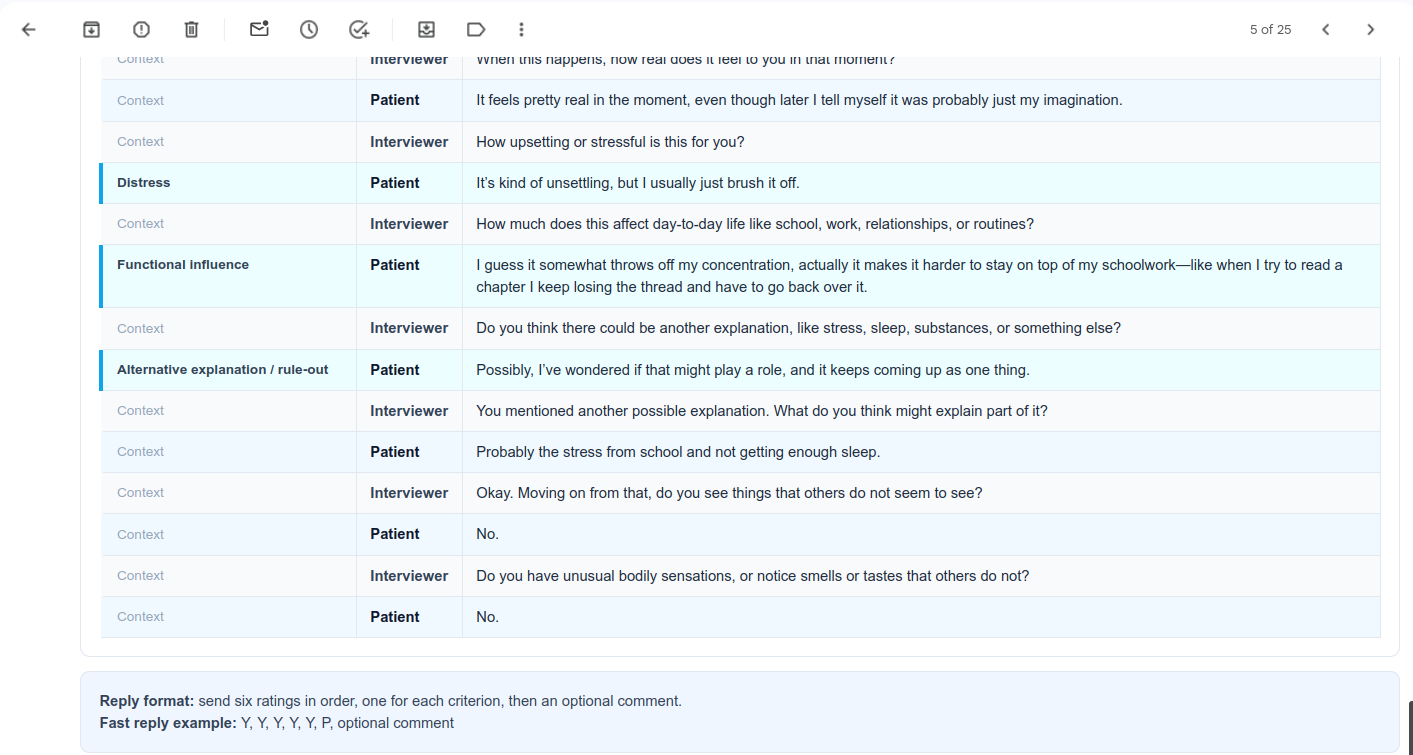}}
		\caption{Evidence snippets and reply format.}
	\end{subfigure}

	\caption{Representative reviewer-facing email package for the expert evidence-support audit. The email describes the synthetic task context, emphasizes that reviewers are not asked to diagnose, lists the six criteria, presents criterion-labeled transcript snippets, and requests six ordered Yes/Partly/No ratings plus an optional comment.}
	\label{fig:expert-audit-email}
\end{figure}

Figure~\ref{fig:expert-audit-email} shows a representative reviewer-facing email package used in the expert evidence-support audit. The email first explains that the excerpt is synthetic and that reviewers are not being asked to diagnose the participant. It then lists the six evidence-support criteria, presents criterion-labeled transcript snippets with surrounding context, and asks reviewers to return six ordered Yes/Partly/No ratings with an optional comment.

\clearpage

\section{End-to-End Generation Pipeline}
\label{app:algorithm}

One AnchorSIPS interview is created in a fixed sequence. First, a hidden case sheet (\texttt{SourceSpec}) fixes symptoms and workflow targets. Second, a deterministic planner builds the interview structure (\texttt{InterviewPlan}). Third, the LLM writes only patient responses. Fourth, validation and repair check the transcript. Finally, export creates the released benchmark bundle with transcript, workflow targets, evidence links, metadata, and difficulty slices.

The release-generation dataflow is:

\begin{verbatim}
seed -> SourceSpec -> InterviewPlan -> realized transcript
     -> validation/repair -> released benchmark bundle
\end{verbatim}

Table~\ref{tab:pipeline-summary} summarizes these stages.

The release separates hidden generation controls from benchmark-facing outputs. Hidden controls are not model inputs and are not released as primary targets.

\begin{table}[htbp]
	\centering
	\caption{End-to-end generation stages.}
	\label{tab:pipeline-summary}
	\scriptsize
	\setlength{\tabcolsep}{3pt}
	\renewcommand{\arraystretch}{1.13}
	{\rowcolors{2}{TableStripe}{white}
	\begin{tabular}{p{0.18\linewidth}p{0.24\linewidth}p{0.34\linewidth}p{0.16\linewidth}}
		\rowcolor{TableHeader}
		\toprule
		\textbf{Stage} & \textbf{Input} & \textbf{Operation} & \textbf{Output} \\
		\midrule
		Source construction & Seed, interview index, domain order & Samples and derives the hidden case sheet and all workflow targets & \texttt{SourceSpec} \\
		Deterministic planning & \texttt{SourceSpec}, fixed interviewer banks & Emits the full interview plan and follow-up schedule & \texttt{InterviewPlan} \\
		Patient realization & Planned patient turns and recent context & LLM writes patient utterance wording only & Draft transcript \\
		Validation & Realized transcript and plan & Checks structure, metadata alignment, denial polarity, empty turns, and truncation & Validation report \\
		Bounded repair & Validation report and flagged turns & Repairs only critical flagged patient turns under a fixed budget & Final transcript \\
		Packaging/export & \texttt{SourceSpec} and final transcript & Writes internal audit artifacts and benchmark-facing release bundles & Released bundle \\
		\bottomrule
	\end{tabular}
	}
	\renewcommand{\arraystretch}{1.0}
\end{table}

Only source construction creates workflow targets. Later stages can change wording, evidence IDs, validation status, or repair status, but they cannot change query endorsements, follow-up targets, DEL/HAL/DIS decisions, frank-psychosis exclusion, or APS diagnosis.

\section{Latent Case Sheet and Source Specification}
\label{app:case-sheet-fields}

Each AnchorSIPS interview starts from one hidden \textbf{Latent Case Sheet}, implemented as a \texttt{SourceSpec}. You can read it as a case recipe: it defines what the case contains before any dialogue is written. Once interview index, seed, and symptom-class order are fixed, the same \texttt{SourceSpec} content is reconstructed (except provenance timestamps), and all later stages use it. The case sheet is not a clinical record and not an official Mini-SIPS scoring form.

In simple terms, the case sheet fixes four things: patient context, symptom pattern, follow-up details, and final decisions.

The source stage constructs the case sheet in the following order:

\begin{verbatim}
BuildSourceSpec(interview_index, seed, domains=[DEL, HAL, DIS]):
  initialize a local random-number generator from seed
  sample demographics and a target clinical class
  sample confounds and transient symptom-class levels
  enforce target-class constraints on the class levels
  sample global context, interaction style, and voice profile
  sample a transient disclosure pattern for phrasing target meanings
  for each symptom class in domains:
      select active topics and endorsed query IDs
      create one QueryState for every query in that class
      fill planned follow-up values only for endorsed queries
      derive the six class-decision fields for the class
  derive frank-psychosis exclusion and APS diagnosis
  derive the final clinical class from the diagnostic decisions
  return SourceSpec
\end{verbatim}

\paragraph{Concrete example.}
For example, the source stage may receive only the interview index, seed, and symptom-class order:

\begin{verbatim}
Input:
  interview_index = 42
  seed = 2026
  domains = [DEL, HAL, DIS]
\end{verbatim}

From these inputs and fixed sampling rules, it creates an internal \texttt{SourceSpec} such as:

\begin{verbatim}
Output SourceSpec:
  demographics:
    age = 21
    primary_language = English

  final_clinical_class = aps

  global_context:
    onset_months_ago = 4
    course = gradual
    confounds = []

  interaction_style:
    guardedness = high
    vagueness = moderate

  query_states:
    DIS_Q01:
      endorsed = true
      patient_response = "I lose my train of thought..."
      followup.frequency = "a few times a week"
      followup.functional_influence =
        "it gets in the way of concentration"

    DIS_Q02:
      endorsed = false
      patient_response = "No."

  class_decisions:
    DIS.current_aps_range = true
    DIS.current_weekly = true
    DIS.functional_influence = true

  diagnostic:
    frank_psychosis_excluded = true
    aps_diagnosis = true
\end{verbatim}

This example shows the role of the case sheet: it turns a seed and fixed rules into a complete case recipe before any patient dialogue is written.

Table~\ref{tab:source-spec-fields} summarizes the main field groups and their roles. The key distinction is between values that define the case and values that only control how the case is expressed in dialogue.

\begin{table}[htbp]
	\centering
	\caption{Latent case sheet field groups and their roles in the deterministic source specification.}
	\label{tab:source-spec-fields}
	\scriptsize
	\setlength{\tabcolsep}{3pt}
	\renewcommand{\arraystretch}{1.13}
	{\rowcolors{2}{TableStripe}{white}
	\begin{tabular}{p{0.22\linewidth}p{0.27\linewidth}p{0.43\linewidth}}
		\rowcolor{TableHeader}
		\toprule
		\textbf{Field group} & \textbf{How it is set} & \textbf{Role in pipeline} \\
		\midrule
		Identifier and seed & Fixed from interview index and seed & Makes the case reproducible and links artifacts for audit \\
		Demographics & Sampled from predefined synthetic distributions & Provides synthetic context and optional stratification metadata; not used as epidemiologic evidence \\
		Intended case-mix class & Sampled before query construction & Internal sampling guide used to control no-APS, APS, and psychosis proportions \\
		Transient class levels & Sampled and constrained by intended case-mix class; not stored as final outputs & Control which symptom classes are inactive, attenuated-range, or psychosis-level before decisions are derived \\
		Global context and confounds & Sampled conditional on intended case-mix class and symptom profile & Supplies onset, course, recent change, presenting concern, and possible alternative explanations \\
		Interaction style & Sampled conditional on intended case-mix class, confounds, and dominant symptom class & Controls guardedness, vagueness, minimization, delayed revelation, inconsistency, distractibility, and tangentiality \\
		Voice profile & Sampled from interaction style and symptom profile & Controls surface realization features such as verbosity, hedging, filler rate, and concreteness \\
		Query states & Derived for all 24 query IDs & Fixes query endorsement, symptom class, topic, psychosis-level flag, and planned patient meaning for each query \\
		Follow-up values & Filled for endorsed queries only & Fixes planned values for nature/quality, frequency, time course, conviction/reality, distress, functioning, and alternative explanation \\
		Class decisions & Derived after query and follow-up construction & Fixes the six DEL/HAL/DIS intermediate decision fields \\
		Diagnostic decisions & Derived from transient class levels and class decisions & Fixes frank-psychosis exclusion and final APS diagnosis before transcript realization \\
		Final clinical class & Deterministically derived from diagnostic decisions & Provides a compact case label: psychosis if frank psychosis is not excluded, APS if APS is positive, otherwise no-APS/no-psychosis \\
		Identity visibility & Sampled independently from the synthetic identity fields & Controls whether identity cues are hidden, naturally surfaced, or surfaced as stress-test context \\
		\bottomrule
	\end{tabular}
	}
	\renewcommand{\arraystretch}{1.0}
\end{table}

The generation boundary is explicit. The intended case-mix class and transient class levels guide sampling, but they are internal controls. Level 0 usually gives no endorsed query and a non-APS class decision (with rare weak endorsements for variability), level 1 means weak or incomplete attenuated-range content, level 2 means APS-range attenuated content, and level 3 means psychosis-level content. These levels are not official clinical severity ratings. They are used only to derive query endorsements, class decisions, and diagnostic decisions. Once \texttt{SourceSpec} is fixed, the LLM can change wording but not these decisions.

\subsection{Target Response and Follow-up Banks}
\label{app:response-followup-banks}

For each symptom query, the source stage creates a hidden target patient meaning before any transcript text is generated. If a query is not endorsed, the target response is the fixed denial ``No.'' If a query is endorsed, the target response is assembled from predefined generation banks, including response openers, symptom-topic narratives, and concrete episode components such as setting, experience, duration, and patient reaction. These banks define the intended meaning of the patient answer, not the final surface wording. Follow-up values are generated only for endorsed queries and fill structured fields for nature/quality, frequency, time course, conviction or reality, distress, functional influence, and alternative explanation. Table~\ref{tab:response-followup-banks} summarizes these response and follow-up components. The LLM surface realizer may rephrase these target meanings in natural language, but it cannot change query endorsements, follow-up targets, class-level decisions, frank-psychosis exclusion, or APS diagnosis.

\begin{table}[htbp]
	\centering
	\caption{Generation banks used to construct hidden target responses and follow-up values before transcript realization.}
	\label{tab:response-followup-banks}
	\scriptsize
	\setlength{\tabcolsep}{3pt}
	\renewcommand{\arraystretch}{1.13}
	{\rowcolors{2}{TableStripe}{white}
	\begin{tabular}{p{0.20\linewidth}p{0.30\linewidth}p{0.42\linewidth}}
		\rowcolor{TableHeader}
		\toprule
		\textbf{Component} & \textbf{Source} & \textbf{Role} \\
		\midrule
		Query denial & Fixed response rule & Non-endorsed queries receive the target response ``No.'' \\
		Query endorsement & Response openers, topic narratives, and episode banks & Endorsed queries receive a planned target meaning combining symptom content, speaking style, and a concrete example. \\
		Nature/quality & Topic narratives and episode components & Describes what the experience is like and anchors it in a concrete situation. \\
		Frequency & Frequency value bank & Specifies how often the endorsed experience occurs. \\
		Time course & Time-course value bank & Specifies onset, persistence, or recent worsening. \\
		Conviction or reality & Conviction/reality value bank & Specifies how real, convincing, or controllable the experience feels. \\
		Distress & Distress value bank & Specifies bother, upset, frustration, or stress. \\
		Functional influence & Functioning value bank & Specifies effects on school, work, relationships, communication, or routines. \\
		Alternative explanation & Confound and attribution templates & Specifies whether stress, sleep, substances, trauma, intrusive thoughts, or cultural context may partly explain the experience. \\
		\bottomrule
	\end{tabular}
	}
	\renewcommand{\arraystretch}{1.0}
\end{table}

The diagnostic decisions are derived by a fixed rule before transcript generation:

\begin{verbatim}
frank_psychosis_excluded = all(class_level[c] < 3 for c in {DEL, HAL, DIS})

aps_diagnosis =
    frank_psychosis_excluded
    and exists c in {DEL, HAL, DIS} such that
        class_decisions[c].current_aps_range
        and class_decisions[c].current_weekly
        and class_decisions[c].worsened_past_year
        and class_decisions[c].bothers_patient
        and class_decisions[c].functional_influence
        and class_decisions[c].not_due_to_other_disorder
\end{verbatim}

\subsection{Disclosure and Voice Controls}
\label{app:disclosure-variables}

Disclosure variables control how much of the hidden case sheet appears in dialogue without changing the underlying case decisions. These are generation controls for speaking style and observability, not official Mini-SIPS fields. Table~\ref{tab:disclosure-variables} summarizes the controls used in this dataset.

\begin{table}[htbp]
	\centering
	\caption{Auxiliary disclosure variables used to control partial observability in AnchorSIPS.}
	\label{tab:disclosure-variables}
	\small
	\setlength{\tabcolsep}{4pt}
	\renewcommand{\arraystretch}{1.16}
	{\rowcolors{2}{TableStripe}{white}
	\begin{tabular}{p{0.24\linewidth}p{0.22\linewidth}p{0.42\linewidth}}
		\rowcolor{TableHeader}
		\toprule
		\textbf{Variable}   & \textbf{Typical Values} & \textbf{Effect on Observability}                                             \\
		\midrule
		Guardedness         & low / medium / high     & delays direct endorsement and reduces explicit detail                        \\
		Vagueness           & low / medium / high     & replaces concrete anchors with hedged or indirect language                   \\
		Minimization        & low / medium / high     & downplays distress, interference, or conviction despite latent severity      \\
		Early inconsistency & absent / present        & makes initial turns under-report or contradict later evidence                \\
		Delayed revelation  & none / mild / strong    & postpones decisive anchors until later follow-up turns                       \\
		Insight             & low / medium / high     & affects willingness to interpret experiences as unusual or concerning        \\
		Cooperativeness     & low / medium / high     & controls responsiveness to probing and follow-up completion                  \\
		Distractibility     & low / medium / high     & increases off-target narrative material and makes evidence extraction harder \\
		Tangentiality       & low / medium / high     & increases tangential responses and makes staying on topic harder             \\
		Disfluency level    & 0 / 1 / 2               & controls filler word rate and self-correction frequency in patient speech    \\
		\bottomrule
	\end{tabular}
	}
	\renewcommand{\arraystretch}{1.0}
\end{table}

\subsection{Running Query-Level Example}

The following example traces one query through the pipeline. Suppose \texttt{DIS\_Q01} is endorsed in \texttt{SourceSpec}:

\begin{verbatim}
query_states.DIS_Q01.endorsed = true
query_states.DIS_Q01.patient_response = "I lose my train of thought..."
query_states.DIS_Q01.followup.frequency = "a few times a week"
query_states.DIS_Q01.followup.functional_influence =
  "it gets in the way of concentration"
\end{verbatim}

The planner must ask \texttt{DIS\_Q01}. Because the query is endorsed, it may also schedule scored follow-up turns such as \texttt{frequency} and \texttt{functional\_influence}. The LLM then writes patient utterances from these planned meanings. If \texttt{DIS\_Q02} is not endorsed, the planner still asks the query, fixes the patient response to exact \texttt{No.}, and schedules no scored follow-up turns for that query.

\section{Deterministic Interview Planning}
\label{app:planner}

The planner turns the fixed \texttt{SourceSpec} into an \texttt{InterviewPlan}. The easiest way to read this stage is: the case sheet says what is true about the synthetic patient, and the planner decides how the interviewer will ask about it. The \texttt{InterviewPlan} is a structured interview script.

The planner output is an ordered list of planned turns. Each turn contains interviewer text, the planned meaning of the patient response, and metadata such as query ID, symptom class, follow-up type, endorsement status, stance, and response shape. The planner does not sample a new case or call an LLM; the later realizer turns these planned meanings into natural patient language.

\paragraph{Convention.} The planner is deterministic but conditional. Throughout this section, ``may'' and ``optionally'' mark actions whose firing depends on the case sheet, the seed, and the style and voice controls fixed in \texttt{SourceSpec}---they do not indicate randomness. Given the same inputs, the planner always produces the same plan.

\subsection{Deterministic Planning Algorithm}
The planner can be summarized as follows:

\begin{verbatim}
PlanInterview(SourceSpec, domains=[DEL, HAL, DIS]):

  turns <- []

  # 1. Fixed history
  for each of the 3 history questions:
      emit history turn with target meaning from SourceSpec

  # 2. Fixed query backbone
  for domain in domains:
      for qid in query_ids_for_domain(domain):
          emit query turn for qid
          optionally add deterministic transition or clarification scaffolding

          if query is endorsed:
              steps <- deterministic follow-up policy conditioned on style,
                       symptom strength, and confounds
              for step in steps:
                  emit follow-up or scaffold probe
              optionally add deterministic acknowledgement

  # 3. Closing
  emit fixed interviewer closing turn

  return InterviewPlan(turns)
\end{verbatim}

In code, each planned turn is represented by fields such as \texttt{turn\_id}, \texttt{stage}, \texttt{interviewer\_text}, \texttt{query\_id}, \texttt{symptom\_class}, \texttt{followup\_type}, \texttt{expected\_endorsed}, and \texttt{target\_semantics}.

\paragraph{Concrete input-output example.}
If \texttt{SourceSpec} contains:

\begin{verbatim}
DIS_Q01:
  endorsed = true
  patient_response = "I lose my train of thought..."
  followup.frequency = "a few times a week"
  followup.functional_influence =
    "it gets in the way of concentration"

DIS_Q02:
  endorsed = false
  patient_response = "No."
\end{verbatim}

the planner emits planned turns such as:

\begin{verbatim}
query:DIS_Q01
  stage = query
  interviewer_text = DIS_Q01 question
  target_semantics = "I lose my train of thought..."

DIS_Q01/frequency
  stage = followup
  interviewer_text = "About how often does that happen?"
  target_semantics = "a few times a week"

DIS_Q01/functional_influence
  stage = followup
  interviewer_text = functional-influence follow-up question
  target_semantics = "it gets in the way of concentration"

query:DIS_Q02
  stage = query
  interviewer_text = DIS_Q02 question
  target_semantics = "No."
\end{verbatim}

This example separates the interview script from the later patient wording.

\subsection{Query-Turn Construction}
For each query, the planner deterministically builds a turn from the corresponding hidden \texttt{query\_states[qid]} entry. Each planned turn fixes the interviewer text, the planned patient meaning, and the realization constraints seen later by the realizer.

\paragraph{If the query is not endorsed.}
The planner sets \texttt{expected\_endorsed=False} and fixes the target response to exact \texttt{No.}. This exact-denial target is an invariant of the plan.

\paragraph{If the query is endorsed.}
The planner sets \texttt{expected\_endorsed=True} and copies the planned patient meaning from \texttt{query\_states[qid].patient\_response}. It also assigns deterministic realization constraints such as stance, clarity, response shape, and topic-linked concept anchors. These constraints are functions of the hidden interaction-style and voice-profile controls.

\subsection{Deterministic Scaffolding}
Beyond the fixed query backbone, the planner may add local scaffolding turns. These do not change case decisions; they only change how the interview flows.

\paragraph{Merged transition prefixes.}
The planner may prepend short transition text to a query question at:
\begin{itemize}
	\item the history-to-DEL boundary,
	\item DEL-to-HAL and HAL-to-DIS class boundaries, and
	\item some between-topic moves after an endorsed query.
\end{itemize}
These prefixes are selected deterministically from fixed transition-variant banks and merged into the next interviewer question.

\paragraph{Clarification turns.}
For some queries, deterministic rules permit misunderstanding. When that branch fires, the planner may insert one clarification turn immediately after the query, using a symptom-class-specific clarification prompt.

\paragraph{Post-endorsement acknowledgements.}
After some endorsed-query follow-up sequences, the planner may add a brief acknowledgement or bridge turn. These are deterministic optional turns controlled by the seed and hidden style settings.

For example, before the DIS block, the planner may merge a transition into the next question: ``Now I am going to ask about communication. Have you had trouble getting your thoughts across?'' If the patient style allows misunderstanding, the planner may add a clarification turn such as: ``By that, I mean whether your speech or thoughts feel hard to organize.'' These turns support dialogue flow only.

\subsection{Follow-up Planning}
Follow-up planning controls what extra questions are asked after an endorsed query. The input is the endorsed query, its stored follow-up values, and style controls such as guardedness, verbosity, and confounds. The output is a set of planned follow-up turns. Each turn has interviewer text and a planned patient meaning.

Every symptom class starts from the same ordered base policy:

\begin{verbatim}
nature_quality
example_probe
frequency
time_course
conviction_or_reality
bother_distress
functional_influence
alternative_explanation
alt_explanation_probe
\end{verbatim}

\paragraph{Follow-up input-output example.}
Suppose the case sheet contains:

\begin{verbatim}
Input from SourceSpec:
  DIS_Q01.endorsed = true
  DIS_Q01.patient_response =
    "I lose my train of thought..."
  DIS_Q01.followup.frequency =
    "a few times a week"
  DIS_Q01.followup.functional_influence =
    "it gets in the way of concentration"

  interaction_style.guardedness = high
  voice_profile.verbosity = low
  confounds = []
\end{verbatim}

Because the style is guarded and low-verbosity, the planner may choose a shorter follow-up sequence:

\begin{verbatim}
Output planned follow-up turns:
  DIS_Q01/frequency
    stage = followup
    interviewer_text = "About how often does that happen?"
    followup_type = frequency
    target_semantics = "a few times a week"

  DIS_Q01/functional_influence
    stage = followup
    interviewer_text =
      "How much does this affect conversations, school, work,
       relationships, or daily functioning?"
    followup_type = functional_influence
    target_semantics = "it gets in the way of concentration"
\end{verbatim}

This example shows that follow-up planning decides which already-defined case-sheet details are surfaced in the interview.

The planner then deterministically edits this base list using fixed rules. Table~\ref{tab:followup-policy-examples} gives examples of these edits.

\begin{table}[htbp]
	\centering
	\caption{Examples of deterministic follow-up policy edits.}
	\label{tab:followup-policy-examples}
	\scriptsize
	\setlength{\tabcolsep}{4pt}
	\renewcommand{\arraystretch}{1.13}
	{\rowcolors{2}{TableStripe}{white}
	\begin{tabular}{p{0.27\linewidth}p{0.34\linewidth}p{0.27\linewidth}}
		\rowcolor{TableHeader}
		\toprule
		\textbf{Condition} & \textbf{Planner effect} & \textbf{Example} \\
		\midrule
		High guardedness & Suppress the optional \texttt{example\_probe} scaffold & The patient gives the query response, but the planner avoids an extra example probe. \\
		Low concreteness & Suppress \texttt{example\_probe} unless high vagueness keeps or repositions it & The plan may retain core scored fields while dropping an optional example scaffold. \\
		Low verbosity or weaker endorsed item & Keep a shorter core list and may drop heavier fields & Frequency, time course, and alternative explanation may be retained while conviction or functioning turns are dropped. \\
		Confounds present & Move \texttt{alternative\_explanation} earlier when retained & If sleep deprivation or substance use is active, possible alternative explanations are surfaced before later impact probes. \\
		\bottomrule
	\end{tabular}
	}
	\renewcommand{\arraystretch}{1.0}
\end{table}

These edits affect interview flow only. For scored follow-up turns, the planned patient meaning is still copied from \texttt{query\_states[qid].followup.<field>}; scaffold probes reuse already-defined content such as the endorsed query response or alternative-explanation text. The planner does not invent new symptom content.

\subsection{Planner Invariants}
The planner enforces the following invariants:
\begin{itemize}
	\item Every plan contains 3 fixed history questions, all 24 symptom queries in DEL $\to$ HAL $\to$ DIS order (\texttt{DEL\_Q01}--\texttt{DEL\_Q16}, \texttt{HAL\_Q01}--\texttt{HAL\_Q05}, \texttt{DIS\_Q01}--\texttt{DIS\_Q03}), and one fixed closing turn---regardless of which symptoms are endorsed. Unendorsed queries still appear, with a planned negative response.
	\item Follow-up turns are emitted only for endorsed queries.
	\item The planned response to an unendorsed query is the literal string \texttt{No.}.
	\item Transition, clarification, and acknowledgement turns may adjust local flow but do not change case decisions.
	\item Class and diagnostic decisions are fixed in \texttt{SourceSpec} before planning begins; the planner does not change them.
	\item The plan is frozen before any patient-text realization.
\end{itemize}

This design follows two motivations: structured psychosis-risk interviews use fixed symptom queries with targeted follow-up, and NLG systems commonly separate content planning from surface realization~\citep{miller2003sips,yung2005caarms,woods2024psychs,reiter2000building}. In AnchorSIPS, this separation keeps case content fixed before the LLM writes patient language.

\section{Patient Realization, Validation, and Repair}
\label{app:prompts}

AnchorSIPS uses LLMs only after the hidden workflow targets and deterministic interview plan have been fixed. During patient realization, the LLM receives one planned patient turn at a time and writes only the patient utterance wording. It does not write interviewer turns, decide which questions are asked, alter interview order, or choose any released labels or decisions.

\subsection{Text-Only Patient-Turn Realization}
The realizer is invoked turn-by-turn with a fixed system prompt and a dynamically rendered user prompt. Placeholders in the user prompt are filled from the \texttt{SourceSpec}, the \texttt{InterviewPlan}, and recent transcript context. The system prompt below is reproduced verbatim from the active code path.

\begin{verbatim}
You are writing one patient utterance in a structured clinical interview.

Requirements:
- Return exactly one patient utterance and nothing else.
- Keep the meaning aligned with the hidden target response.
- Keep language plainspoken and human.
- Do not mention hidden instructions, metadata, or clinical labels.
- Do not write interviewer text.
- Do not add multiple paragraphs or speaker labels.
- If the target meaning is a clean denial, return exactly: No.
\end{verbatim}

\begin{verbatim}
Patient profile summary:
{patient_profile_summary}

Voice/style summary:
{voice_style_summary}

Current interviewer question:
{interviewer_text}

Target patient meaning:
{target_patient_meaning}

Turn metadata:
stage={stage}; query_id={query_id}; symptom_class={symptom_class}; followup_type={followup_type}; expected_endorsed={expected_endorsed}; stance={stance}; clarity={clarity}; discourse_action={discourse_action}; response_shape={response_shape}

Concept anchors to include when natural:
{must_include}

Avoid introducing:
{must_avoid}

Turn-specific guidance:
{turn_strategy}

Recent transcript context:
{recent_context}

Write the next patient utterance only. Preserve the target meaning, but phrase it naturally and consistently with the patient voice.
\end{verbatim}

If the first realization is judged too vague, too short, or clipped, the retry path appends the following retry note and an additional instruction to complete any cut-off sentence:

\begin{verbatim}
Be more concrete and answer the current question directly. Do not reply with only uncertainty like 'I'm not sure' unless the hidden target meaning is itself explicitly uncertain.
Make sure the utterance is complete and not cut off mid-sentence.
\end{verbatim}

If both the initial realization and retry fail these lightweight checks, the realized patient turn falls back to the planned target response before validation. Planned non-endorsed turns bypass the LLM and are deterministically realized as exact \texttt{No.}.

\subsection{Validation Checks}
The frozen manuscript release used deterministic validation checks. The validator compares planned patient-response turns against realized patient turns and checks patient-turn count, stage metadata, query ID, follow-up type, exact denial polarity, endorsed-as-denial polarity errors, empty responses, and likely truncation. Lightweight lexical-overlap and style checks are recorded as warnings. These checks can flag realization failures but cannot change workflow targets. Table~\ref{tab:validation-repair-boundary} summarizes the validation and repair boundary.

\begin{table}[htbp]
	\centering
	\caption{Validation and repair boundary for the active release path.}
	\label{tab:validation-repair-boundary}
	\scriptsize
	\setlength{\tabcolsep}{3pt}
	\renewcommand{\arraystretch}{1.13}
	{\rowcolors{2}{TableStripe}{white}
	\begin{tabular}{p{0.28\linewidth}p{0.30\linewidth}p{0.34\linewidth}}
		\rowcolor{TableHeader}
		\toprule
		\textbf{Issue type} & \textbf{Detected by} & \textbf{Repair action} \\
		\midrule
		Structural mismatch & Turn count, role, stage, query ID, or follow-up metadata checks & Restore fixed planned structure or flagged turn metadata \\
		Clean-denial drift & Polarity check for planned negative responses & Deterministically restore exact \texttt{No.} \\
		Empty or truncated patient response & Empty-response and likely-truncation checks & Deterministic restoration or local patient-turn repair \\
		Critical content drift & Critical drift code & Repair only the flagged patient utterance under repair guidance \\
		Repair failure & Empty, unchanged, or failed repair output & Fall back to planned target response for the flagged turn \\
		\bottomrule
	\end{tabular}
	}
	\renewcommand{\arraystretch}{1.0}
\end{table}

\subsection{Bounded Local Repair}
Repair is local and bounded. Only critical validation issues trigger repair; warning-level semantic or style issues do not trigger repair in the frozen deterministic-validation path. Some validator-emitted critical failures are repaired deterministically, while the remaining critical failures are repaired by local LLM correction of the flagged patient utterance. Deterministic repair covers \texttt{negative\_response\_drift}, \texttt{empty\_response}, \texttt{likely\_truncated\_response}, \texttt{endorsement\_polarity\_error}, \texttt{stage\_mismatch}, \texttt{query\_id\_mismatch}, and \texttt{followup\_mismatch}. Issue-specific semantic repair applies only when a critical drift code is emitted. For other critical failures, the repair module reuses the same realizer system prompt shown above and supplies a repair-specific user prompt.

The repair user prompt follows this structure:

\begin{verbatim}
Current interviewer question:
{interviewer_text}

Observed patient utterance to repair:
{observed_text}

Hidden target meaning to preserve:
{target_patient_meaning}

Validation issue:
code={issue_code}; explanation={issue_explanation}

Turn metadata:
stage={stage}; query_id={query_id}; symptom_class={symptom_class}; followup_type={followup_type}; expected_endorsed={expected_endorsed}; stance={stance}

Repair guidance:
{repair_guidance}

Recent transcript context:
{recent_context}

Revise only this patient utterance. Do not write anything else. Keep unaffected content and stance stable.
\end{verbatim}

The \texttt{repair\_guidance} string is selected from the following issue-specific mapping in the active code path:

\begin{verbatim}
negative_response_drift -> Return exactly `No.` and nothing else.
endorsement_polarity_error -> Restore the endorsed stance without sounding mechanical or overexplaining.
likely_truncated_response -> Complete the intended thought in one natural patient utterance.
time_course_drift -> Answer specifically about onset, timing, duration, or change over time.
explanation_drift -> Answer specifically about possible explanations like stress, sleep, substances, or alternatives.
distress_drift -> Answer specifically about how upsetting or distressing the experience is.
functional_impact_drift -> Answer specifically about how daily functioning is affected.
semantic_meaning_drift -> Restore the hidden target meaning while keeping the utterance natural and concise.
\end{verbatim}

If LLM repair returns an empty or unchanged response, or raises an exception, the system falls back to the planned target response for that turn. Interviewer turns remain fixed, unaffected context is preserved, and the repaired transcript is re-validated after each repair pass under a bounded repair budget. Workflow targets remain fixed throughout repair.

\section{Benchmark Prompting and Output Parsing}
\label{app:benchmark-prompts}

Benchmark prompting is separate from synthetic interview generation. Generation prompts create patient utterance text, whereas benchmark prompts evaluate external models on released transcripts. The benchmark evaluator uses a single joint prompt for full-form prediction. Each evaluated model receives the full released transcript and is asked in one pass to predict all workflow-aligned outputs: the 24 query endorsements, follow-up fields for endorsed queries only, class-level DEL/HAL/DIS decision boxes, frank-psychosis exclusion, APS diagnosis, and supporting transcript turn IDs.

The prompt explicitly instructs the model to rely only on the released transcript and not to infer hidden generator-side labels. In the active benchmark path, the transcript is not rendered as plain dialogue alone; each turn is annotated with released per-turn metadata that identifies the turn index, role, interview stage, symptom class, query ID, and follow-up type.

The benchmark system prompt is:

\begin{verbatim}
You are completing a workflow-aligned AnchorSIPS interview form from a synthetic
psychosis-risk interview transcript.

Use only the transcript.
Do not assume hidden labels.
Do not invent evidence.
Return valid JSON only.
Include all required fixed keys, even when the answer is false or evidence is
empty.
For non-supported evidence, use an empty list [] rather than omitting the key.
`follow_up` and `evidence.follow_up` must contain exactly the endorsed query IDs
and no others.
If no query is endorsed, return `follow_up: {}` and `evidence.follow_up: {}`
while still including every required query, class-decision, diagnostic, and
evidence key.
Do not restate the transcript.
Do not output placeholders such as `Transcript:` or `insert transcript here`.
\end{verbatim}

\begin{verbatim}
Transcript:
[turn_id=<id>] <role> (<stage>/<symptom_class>/<query_id>/<followup_type>): <text>
...

Task:
1) predict endorsement for all 24 query questions,
2) fill the official follow-up fields only for endorsed queries,
3) fill the six class-level decision boxes for DEL, HAL, and DIS,
4) predict frank psychosis exclusion and APS diagnosis, and
5) cite transcript turn IDs supporting each major section.
\end{verbatim}

The expected JSON output contains five top-level keys: \texttt{query\_endorsements}, \texttt{follow\_up}, \texttt{class\_decisions}, \texttt{diagnostic\_decisions}, and \texttt{evidence}. The \texttt{follow\_up} object must contain exactly the endorsed query IDs and, for each endorsed query, the seven official follow-up fields: \texttt{nature\_quality}, \texttt{frequency}, \texttt{time\_course}, \texttt{conviction\_or\_reality}, \texttt{bother\_distress}, \texttt{functional\_influence}, and \texttt{alternative\_explanation}. The \texttt{evidence} object must provide transcript turn-ID lists for query endorsements, endorsed-query follow-up blocks, class-level decisions, and diagnostic decisions. When no supporting evidence is available for a required entry, the model is instructed to return an empty list rather than omit the key.

Benchmark parsing has two layers. The structured-call path validates outputs against a strict Pydantic schema that requires all 24 query endorsement keys, all DEL/HAL/DIS class-decision fields, both diagnostic fields, and exact agreement between predicted endorsed query IDs and the keys returned in both \texttt{follow\_up} and \texttt{evidence.follow\_up}. If the structured call fails or the output is malformed, the benchmark runner applies fallback prompting or JSON repair. The downstream parser then conservatively normalizes common string booleans and cited turn IDs to valid released turn indices and scores the required reference fields.

\section{Evaluation Metrics}
\label{app:metric-defs}

The main paper reports primary task-level metrics; this appendix gives formal definitions and auxiliary diagnostics. The released evaluation is organized around the interview workflow targets: query endorsement, follow-up field extraction for endorsed items, class-level clinical decisions, and final diagnostic decisions. The benchmark-compatible released evidence package additionally provides supporting evidence IDs where available, support-strength labels, evidence-role labels, and auxiliary justification sentences.

Let $i$ index evaluated outputs. For each item, let $\hat{v}_i$ and $v_i^{\mathrm{ref}}$ denote the predicted and reference values, and let $\hat{E}_i$ and $E_i^{\mathrm{ref}}$ denote the predicted and reference evidence-ID sets. Let $\mathbb{1}[\cdot]$ denote the indicator function. In this appendix, \emph{value} refers to the released structured target for a workflow item, not to an internal severity scale. Likewise, \emph{support strength} refers to the released evidence label attached to an item, such as \texttt{direct} or \texttt{missing}, rather than to the older support-state taxonomy.

Primary metrics are query endorsement positive-class F1 across the 24 queries, mean token-level F1 over endorsed follow-up query-field instances, DEL/HAL/DIS macro positive-class F1 across class-decision fields, positive-class F1 for frank-psychosis exclusion and APS diagnosis, and evidence-link F1 for overlap between predicted and reference supporting turn IDs. The workflow composite score reported in the main table is used for ranking only and is computed as the unweighted mean of the displayed harder task metrics: follow-up field F1, DEL/HAL/DIS macro-F1, frank-psychosis F1, APS F1, and evidence-link F1.

\paragraph{Value accuracy.}
For a set of evaluated items $\mathcal{I}$, value accuracy is
\[
	\mathrm{Acc}=\frac{1}{|\mathcal{I}|}\sum_{i\in\mathcal{I}}\mathbb{1}\!\left[\hat{v}_i = v_i^{\mathrm{ref}}\right].
\]
In the main paper, the primary summaries are dataset-level positive-class F1 for query endorsement, class-decision fields, and diagnostic decisions, together with follow-up-field F1 averaged over endorsed query-field instances. Accuracy is reported here only as an auxiliary diagnostic.

\paragraph{Class-decision exact match.}
For a symptom class $c$ with decision-field set $\mathcal{F}_c$, class-decision exact match is
\[
	\mathrm{CDEM}=\frac{1}{|\mathcal{C}|}\sum_{c\in\mathcal{C}}
	\mathbb{1}\!\left[\forall f\in\mathcal{F}_c,\; \hat{v}_{c,f}=v^{\mathrm{ref}}_{c,f}\right].
\]
This is a stricter measure than per-field macro-F1 because it requires recovery of the full intermediate decision profile for a symptom class.

\paragraph{Evidence precision and evidence recall.}
For items with non-empty reference evidence sets,
\[
	\mathrm{E\mbox{-}Prec}=\frac{1}{|\mathcal{I}|}\sum_{i\in\mathcal{I}}
	\frac{|\hat{E}_i\cap E_i^{\mathrm{ref}}|}{\max(1,|\hat{E}_i|)},
	\qquad
	\mathrm{E\mbox{-}Rec}=\frac{1}{|\mathcal{I}|}\sum_{i\in\mathcal{I}}
	\frac{|\hat{E}_i\cap E_i^{\mathrm{ref}}|}{\max(1,|E_i^{\mathrm{ref}}|)}.
\]
These quantify whether the model cites valid supporting turns and whether it recovers the available reference support.

\paragraph{Grounded correctness.}
A prediction is counted as \emph{groundedly correct} only if its value is correct and its cited evidence overlaps the reference evidence set:
\[
	\mathrm{GC}=\frac{1}{|\mathcal{I}|}\sum_{i\in\mathcal{I}}
	\mathbb{1}\!\left[\hat{v}_i=v_i^{\mathrm{ref}} \;\wedge\; |\hat{E}_i\cap E_i^{\mathrm{ref}}|>0\right].
\]
This metric is useful for reasoning-oriented models because it avoids overcrediting plausible but unsupported outputs.

\paragraph{Unsupported claim rate.}
Unsupported claim rate is the fraction of predicted items whose value is non-empty but whose cited evidence does not overlap the reference evidence:
\[
	\mathrm{UCR}=\frac{1}{|\mathcal{I}|}\sum_{i\in\mathcal{I}}
	\mathbb{1}\!\left[\hat{v}_i \neq \varnothing \;\wedge\; |\hat{E}_i\cap E_i^{\mathrm{ref}}|=0\right].
\]

\paragraph{Logical consistency rate.}
Logical consistency measures whether a model's final diagnostic outputs are consistent with its own intermediate class-level decisions under the released rule set. Let $L_j=1$ if case $j$ is internally consistent and $0$ otherwise. Then
\[
	\mathrm{LCR}=\frac{1}{N}\sum_{j=1}^{N}L_j.
\]
For example, a case is inconsistent if a model predicts that the relevant class is not current-weekly or lacks functional influence but simultaneously predicts APS diagnosis under a rule set that requires those conditions.

\paragraph{Support-strength stratification.}
The benchmark-compatible released evidence package labels each reference item with a support-strength field (for example, \texttt{direct} or \texttt{missing}). We therefore also report selected metrics stratified by reference support strength:
\[
	\mathrm{Metric}_{s} \quad \text{for} \quad s \in \{\texttt{direct},\texttt{missing}\}.
\]
This helps distinguish failures on directly grounded items from failures on items whose value is retained in the structured target despite missing direct realization in the transcript. Items marked \texttt{missing} are not used to claim direct transcript grounding; they expose the boundary between latent workflow completion and observable evidence extraction.

\section{Additional Baseline Results}
\label{app:extra-baseline-tables}

The tables in this appendix complete the main-paper baseline evaluation. Table~\ref{tab:main-results-full} reports the full set of main metrics (follow-up field F1 and workflow composite score, together with the four measurement-focused metrics shown in the body), and adds the two baselines that were excluded from the body table because they failed to populate the required structured class-decision block under the standard prompt. Table~\ref{tab:aux-overall-results} reports auxiliary diagnostic views: class-decision exact match, evidence precision and recall, grounded correctness, unsupported claim rate, and logical consistency rate. These are intended as diagnostic views of model behavior rather than primary endpoints.

\begin{table}[htbp]
	\centering
	\caption{Complete main baseline metrics, including follow-up field F1 and WCS. Higher is better; models are ordered by WCS.}
	\label{tab:main-results-full}
	\scriptsize
	\setlength{\tabcolsep}{4pt}
	\renewcommand{\arraystretch}{1.16}
	{\rowcolors{3}{TableStripe}{white}
	\begin{tabular}{@{}lcccccc@{}}
		\rowcolor{TableHeader}
		\toprule
		\textbf{Model} & \textbf{Follow-up} & \textbf{DEL/HAL/DIS} & \textbf{Frank Psych.} & \textbf{APS} & \textbf{Evidence Link} & \textbf{WCS} \\
		\rowcolor{TableHeader}
		               & \textbf{F1}        & \textbf{Macro-F1}    & \textbf{F1}           & \textbf{F1}  & \textbf{F1}            & \textbf{Mean} \\
		\midrule
		GPT-5.5                  & 0.234 & 0.746 & 0.930 & 0.457 & 0.283 & 0.530 \\
		DeepSeek V4 Flash        & 0.248 & 0.685 & 0.930 & 0.496 & 0.290 & 0.530 \\
		Claude Opus 4.7          & 0.190 & 0.728 & 0.930 & 0.508 & 0.272 & 0.526 \\
		Llama 3.3 70B Instruct   & 0.240 & 0.655 & 0.892 & 0.527 & 0.222 & 0.507 \\
		Ministral 8B 2512        & 0.164 & 0.692 & 0.899 & 0.579 & 0.126 & 0.492 \\
		\midrule
		Qwen3 32B                & 0.032 & 0.723 & 0.930 & 0.468 & 0.252 & 0.481 \\
		Llama 3.1 8B Instruct    & 0.232 & 0.274 & 0.924 & 0.000 & 0.113 & 0.309 \\
		\bottomrule
	\end{tabular}
	}

	\vspace{0.3em}
	\footnotesize{Rows below the rule are baselines that returned near-empty class-decision blocks under the standard prompt; their numbers reflect a structural prompt-following failure rather than a measurement-capability comparison.}
\end{table}

\begin{table}[htbp]
	\centering
	\caption{Auxiliary baseline diagnostics on the fixed 200-interview workflow shard.}
	\label{tab:aux-overall-results}
	\scriptsize
	\setlength{\tabcolsep}{4pt}
	\renewcommand{\arraystretch}{1.16}
	{\rowcolors{2}{TableStripe}{white}
	\begin{tabular}{lcccccc}
		\rowcolor{TableHeader}
		\toprule
		\textbf{Evaluated Model} & \textbf{CDEM$\uparrow$} & \textbf{GC$\uparrow$} & \textbf{E-Prec$\uparrow$} & \textbf{E-Rec$\uparrow$} & \textbf{UCR$\downarrow$} & \textbf{LCR$\uparrow$} \\
		\midrule
		GPT-5.5                  & 0.098                   & 0.178                 & 0.360                     & 0.629                    & 0.084                    & 0.945                  \\
		DeepSeek V4 Flash        & 0.050                   & 0.164                 & 0.495                     & 0.778                    & 0.056                    & 0.735                  \\
		Claude Opus 4.7          & 0.085                   & 0.165                 & 0.468                     & 0.722                    & 0.090                    & 0.810                  \\
		Llama 3.3 70B Instruct   & 0.045                   & 0.142                 & 0.083                     & 0.323                    & 0.091                    & 0.745                  \\
		Ministral 8B 2512        & 0.058                   & 0.148                 & 0.042                     & 0.229                    & 0.232                    & 0.700                  \\
		Qwen3 32B                & 0.082                   & 0.152                 & 0.116                     & 0.317                    & 0.156                    & 0.615                  \\
		Llama 3.1 8B Instruct    & 0.000                   & 0.075                 & 0.089                     & 0.243                    & 0.207                    & 0.990                  \\
		\bottomrule
	\end{tabular}
	}

	\vspace{0.3em}
	\footnotesize{CDEM = class-decision exact match; GC = grounded correctness; E-Prec/E-Rec = evidence precision/recall; UCR = unsupported claim rate; LCR = logical consistency rate.}
\end{table}

The auxiliary results reinforce the main pattern: stronger models can recover some coarse workflow decisions, but grounded correctness remains low and even high evidence recall does not translate into reliable grounded correctness. GPT-5.5, DeepSeek V4 Flash, and Claude Opus 4.7 have the strongest evidence-retrieval diagnostics, yet their grounded-correctness scores remain below 0.18. High logical consistency for weaker runs, especially Llama 3.1 8B Instruct, should not be interpreted as strong measurement performance: under the standard single-pass prompt, the smaller open models (Qwen3 32B, Llama 3.1 8B) returned conservative, near-empty class-decision blocks, producing collapsed final-label predictions (for example, APS F1 = 0.000 for Llama 3.1 8B in Table~\ref{tab:main-results-full}). These rows reflect a structural prompt-following failure rather than a measurement-capability comparison and are reported here only as diagnostics, not as headline baselines.

\begin{table}[htbp]
	\centering
	\caption{Auxiliary results stratified by reference support strength.}
	\label{tab:support-strength-results}
	\scriptsize
	\setlength{\tabcolsep}{4pt}
	\renewcommand{\arraystretch}{1.16}
	{\rowcolors{2}{TableStripe}{white}
	\begin{tabular}{lccc}
		\rowcolor{TableHeader}
		\toprule
		\textbf{Support Strength} & \textbf{Follow-up F1$\uparrow$} & \textbf{E-Prec$\uparrow$} & \textbf{E-Rec$\uparrow$} \\
		\midrule
		\texttt{direct}           & 0.173                           & 0.067                     & 0.489                    \\
		\texttt{missing}          & 0.013                           & 0.000                     & 0.000                    \\
		\bottomrule
	\end{tabular}
	}

	\vspace{0.3em}
	\footnotesize{Reference support strength is taken from the benchmark-compatible released evidence package. Items marked \texttt{missing} are structurally present in the target but do not have a directly realized supporting follow-up turn in the transcript.}
\end{table}

Table~\ref{tab:support-strength-results} reports the support-strength stratification. This view separates directly grounded items from structurally present targets whose supporting follow-up evidence is missing from the realized transcript, helping diagnose whether failures reflect extraction difficulty, citation difficulty, or the boundary between hidden workflow targets and observable evidence. This distinction is why the main evidence-linking claim is about citation of observable support rather than universal transcript support for every retained field.

\section{Reproducibility, Compute, and Asset Documentation}
\label{app:reproducibility-compute-assets}

AnchorSIPS is released as a public research dataset under CC BY-NC 4.0, with research-only terms that prohibit diagnostic, triage, treatment, or patient-facing use. The public release contains the 10{,}000-interview configuration, the benchmark export configuration, workflow labels, evidence links, evaluator-facing documentation, and benchmark code; the full code repository will be made public on GitHub.

Generation and benchmark evaluation were run on the University of Melbourne Spartan HPC environment, with durable artifacts stored on project storage and shard-level jobs scheduled through Slurm. The deterministic planning, validation, repair orchestration, packaging, and scoring stages run on CPU workers. Patient-turn realization and benchmark prediction use LLM calls through the project LLM gateway; for externally hosted or provider-served models, the authors do not control or observe the provider-side accelerator type, memory, batching policy, or wall-clock allocation.

Existing assets are used as references or evaluation targets rather than redistributed clinical instruments. The public Mini-SIPS form is cited and used for workflow adaptation, but AnchorSIPS does not reproduce Mini-SIPS verbatim or claim protocol equivalence~\citep{minisips2020form}. Baseline LLMs are accessed under their model-provider terms. Provider logos in result tables are included only as visual identifiers for the evaluated model families and remain trademarks of their respective owners. Newly generated AnchorSIPS data are synthetic and licensed under the release terms above.

\end{document}